\documentclass{article}

\usepackage{amsmath}
\usepackage{amsfonts}
\usepackage{array}
\usepackage{makecell}
\usepackage{booktabs}
\usepackage{enumitem}
\usepackage{wrapfig}
\usepackage{tcolorbox}
\usepackage{algorithm}
\usepackage[noend]{algpseudocode}
\usepackage{tabularray} % For the tblr environment
\usepackage{tcolorbox}  % For colored boxes and styling
\usepackage{xcolor}     % For text coloring
\usepackage{graphicx} % for including images
\usepackage{caption}  % for caption outside float environments
\usepackage{multirow} 
\usepackage{subcaption}
\usepackage{wrapfig}
\usepackage{xspace}
\usepackage{hyperref}
\usepackage{cleveref}
\usepackage{varwidth}
\usepackage{arydshln}
\usepackage{mathtools}
\usepackage{caption}
\usepackage{mathabx}
\usepackage{xcolor}
\usepackage[preprint]{corl_2026}

\newcommand{\Framename}{\textbf{S}tate \textbf{P}rediction and \textbf{A}daptive \textbf{C}ommand \textbf{E}xecution\xspace}
\newcommand{\framename}{SPACE\xspace}
\newcommand{\algname}{Action Adapter\xspace}
\title{SPACE: Enabling Learning from Cross-Robot Data Toward Generalist Policies}

\author{
  Haeone Lee$^{1}$\thanks{Correspondence to \texttt{haeone.lee@kaist.ac.kr}} \quad
  Byeongguk Jeon$^{1,2}$ \quad
  Suchae Jeong$^{1,2}$ \quad
  Jian Kim$^{3}$ \quad
  Kimin Lee$^{1,2}$ \\[0.3em]
  $^{1}$KAIST \quad
  $^{2}$Config \quad
  $^{3}$Yonsei University
}
\begin{document}
\maketitle

%===============================================================================

\begin{abstract}
In robot learning, scaling training datasets across diverse embodiments and environments has become a dominant paradigm for learning generalizable robot policies. These policies are commonly trained via behavior cloning to imitate actions from pre-collected demonstrations. However, since robot actions are tied to the dynamics of the data collection robot, different robots may require different actions to achieve the same motion. This discrepancy hinders both policy training and deployment across diverse robots.
To address this, we propose using Cartesian state delta as a universal action representation across robots, and introduce \Framename (\textbf{\framename}) framework. \framename handles robot dynamics variation at three levels: across different embodiments, across hardware units of the same embodiment, and within a single robot during operation. 
It consists of two components: (i) a Cartesian state delta policy that predicts geometric end-effector displacement, and (ii) \algname, which converts the predicted Cartesian state delta into robot-specific control commands. Experiments show that \framename substantially outperforms policies that directly predict control commands when learning from data collected across different embodiments and across hardware units of the same embodiment. \framename also remains robust under dynamics shifts at deployment, including changes in control frequency, object weight, and controller gains. 
The project page is available at \texttt{http://haeone.site/space-website/}.
% The project website is available at \href{http://rl-max.github.io/space-website/}{the project page}.
% Videos are available at \href{http://rl-max.github.io/space-website/}{the project page}.
% In robot learning, scaling training datasets across diverse embodiments and environments has become a dominant paradigm for learning generalizable robot policies. These policies are commonly trained via behavior cloning to imitate actions from pre-collected demonstrations. However, since robot actions are tied to the dynamics of the data collection robot, different robots may require different actions to achieve the same motion. This discrepancy hinders both policy training and deployment across diverse robots.
% To address this, we propose using Cartesian state delta as a universal action representation across robots, and introduce a unified framework that handles dynamics variation at three levels: across different embodiments, across hardware units of the same embodiment, and within a single robot during operation. Our framework consists of two components: (i) a Cartesian state delta policy that predicts geometric end-effector displacement, and (ii) \algname, which converts the predicted Cartesian state delta into robot-specific control commands. Experiments show that our framework substantially outperforms policies that predict control commands when learning from data collected across different embodiments and across hardware units of the same model. Our framework also remains robust under dynamics shifts at deployment, including changes in control frequency, object weight, and controller gains.
\end{abstract}


%===============================================================================
\section{Introduction}
\label{sec:intro}
Foundation models have achieved remarkable success in vision and language by leveraging large-scale, diverse data~\citep{bommasani2021opportunities,
achiam2023gpt, comanici2025gemini}. Recent work extends this paradigm to robotics, training generalizable policies on datasets that span diverse environments and embodiments~\citep{o2024openx, fang2024rh20t, wu2024robomind} or many hardware units of a single embodiment~\citep{khazatsky2024droid, bu2025agibot}. These policies are typically trained via behavior cloning, predicting the actions recorded in the dataset~\citep{o2024openx, kim24openvla, black2024pi0, intelligence2025pi, bjorck2025gr00t}.

A key challenge in leveraging these datasets is that robot actions are inconsistent across them. Different datasets adopt different action spaces (e.g., joint-space versus end-effector commands), and even within a shared action space, the same action produces different motions across robots~\citep{zheng2025universal, bronars2026tune}. The root cause is that actions are defined as input commands to the underlying controller~\citep{khazatsky2024droid, kim24openvla, team2024octo, brohan2023rt}, whose implementation varies across embodiments and even across individual units of the same embodiment.
% due to wear-and-tear and manufacturing variability. 
Furthermore, robot wear-and-tear and manufacturing variability affect the robot motion for the same action.
As a result, a trajectory recorded on one robot is a noisy or even invalid supervision signal for other robots, hindering learning from cross-robot data.

In this work, we present \framename, a unified framework to address this challenge, consisting of two components: (i) a Cartesian state delta policy that predicts end-effector displacement, a controller-independent quantity recoverable directly from recorded trajectories; and (ii) \algname, a lightweight linear model that converts the predicted displacement into a robot-specific control command at deployment time. 
% In this work, we present a unified framework to address this challenge, consisting of two components: (i) a Cartesian state delta policy that predicts end-effector displacement, a controller-independent quantity recoverable directly from recorded trajectories; and (ii) \algname, a lightweight linear model that converts the predicted displacement into a robot-specific control command at deployment time. 
Predicting Cartesian state deltas enables data sharing across robots with different dynamics and kinematics by expressing motion in a shared end-effector space. However, prediction alone is insufficient: executing the predicted displacement on a target robot is non-trivial, since the resulting motion depends on the robot's controller and dynamics. 
% For instance, low stiffness controller gains require larger command than the desired displacement to actually achieve it~\citep{bronars2026tune, kim2025race}. 
% \algname closes this gap, enabling the same policy to be executed on different target robots.
% For instance, a robot with low stiffness gains must be commanded with a larger magnitude than the desired displacement to actually achieve it~\citep{bronars2026tune, kim2025race}. 
\algname closes this gap, enabling the same policy to be executed on different target robots.
% \begin{wrapfigure}{r}{0.48\textwidth}
%     \centering
%     \vspace{-10pt}
%     \includegraphics[width=\linewidth]{figures/figure_space_intro.pdf}
%     \caption{We propose SPACE, a framework that can leverage data collected across different embodiments and hardware using Cartesian state delta as an action.}
%     \label{fig:space_intro}
%     \vspace{-10pt}
% \end{wrapfigure}
\begin{figure}[t]
    \centering
    % \vspace{-5pt}
    % \includegraphics[width=0.7\textwidth]{figures/figure_cross_embodiment_environment.pdf}
    \includegraphics[width=0.95\textwidth]{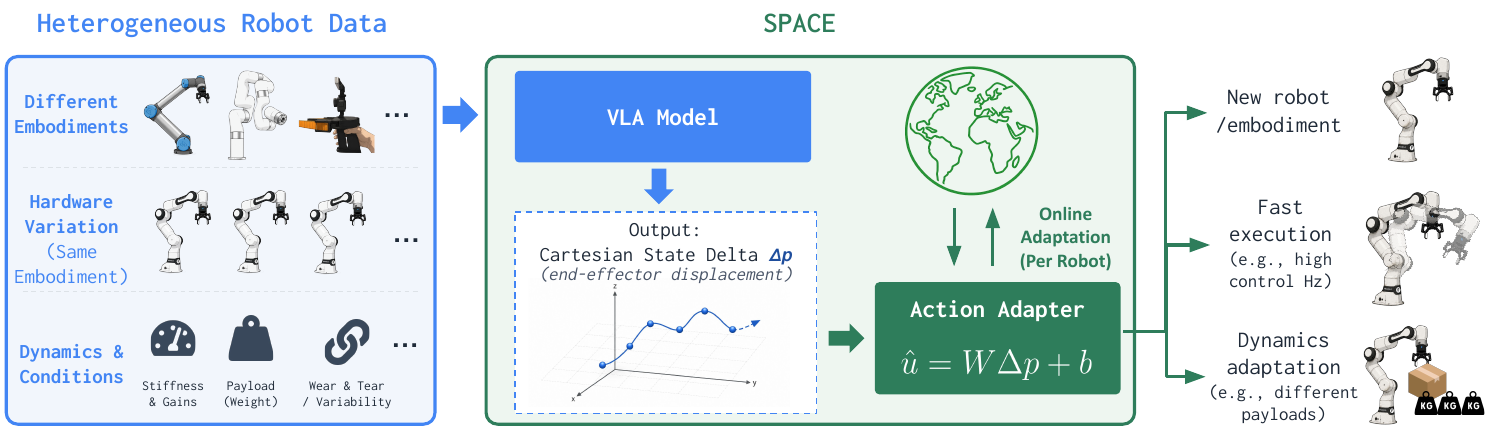}
    \caption{Illustration of SPACE, a framework comprising a Cartesian state-delta policy and \algname. SPACE shares data across embodiments and hardware, supporting deployment on new robots, faster execution (high-frequency control), and dynamics adaptation (different payloads).}
    \label{fig:space_intro}
    \vspace{-15pt}
\end{figure}

We evaluate \framename in three settings. First, in a cross-embodiment setup where a Franka robot policy is co-trained with UR5 data and with human hand-held gripper data, \framename improves success rates over a command-predicting policy by an average of 30\% and 50\%, respectively. Second, when learning from DROID~\citep{khazatsky2024droid}, which spans many hardware units of the same Franka, we achieve an average 54\% gain by absorbing unit-to-unit command discrepancies. Third, we show that \framename remains effective under shifts in control frequency, payload, and controller gains.
These results show that \framename provides a practical recipe for reliable policy learning and deployment across robots.

\section{Related Work}
\label{sec:related_work}
% Domain randomization
% Meta RL
% In-Context Adaptation for Generalizable Imitation Learning
% === REMOVED TO FIT 8 PAGES ===
% \paragraph{Action spaces in robot learning}
% Prior works have explored multiple representations for robot actions, such as commanding joint position~\citep{intelligence2025pi, zhao2023aloha} or end-effector position~\citep{kim24openvla, chi2023diffusion, brohan2023rt, zitkovich2023rt}. Action space also differs in the temporal axis depending on absolute position command or their displacement (delta)~\citep{chi2023diffusion, feng2026demystifying}. Among those, End-effector delta action space has been widely adopted for cross-embodiment learning~\citep{o2024openx, kim24openvla, team2024octo} since it is agnostic to robot-specific kinematics and base-frame translation. Compared to Cartesian state delta in our work, those works use control commands sent to the robot controller as an action, which is coupled with dynamics of data collection robot~\citep{o2024openx, kim24openvla, team2024octo}. 
% Recent works also attempt policy training in latent action, which is sharable across different embodiments~\citep{zheng2025universal, ye2025latent}. However, latent action policy requires fine-tuning to be executed in target robots, while our Cartesian state delta policy is directly executable using \algname.

% === REVISED TO FIT 8 PAGES ===
\textbf{Cross-embodiment policy learning.~}
Several prior works learn policy by combining datasets from different embodiments~\citep{o2024openx, kim24openvla, black2024pi0, intelligence2025pi, bjorck2025gr00t, brohan2023rt, zitkovich2023rt}. 
% Several prior works learn policy by combining datasets from different embodiments~\citep{o2024openx, kim24openvla, brohan2023rt, zitkovich2023rt, bjorck2025gr00t, black2024pi0, intelligence2025pi}.
A few lines of work learn embodiment-specific action heads~\citep{bjorck2025gr00t, team2024octo, wen2025dexvla} or condition embodiment information in the policy~\citep{zheng2025x}. 
Still, this does not fundamentally address the dynamic differences between robots, but separates the effect by training per robot. 
Recent works also attempt policy training in latent action, which is shareable across different embodiments~\citep{zheng2025universal, ye2025latent}. However, the latent action policy requires fine-tuning to be executed in target robots.
% while our Cartesian state delta policy is directly executable using \algname.
% Cosmos Policy~\citep{kim2026cosmos} and DreamZero~\citep{ye2026world} incorporate next frame prediction objective for policy learning, which can be shared across embodiments. However, they do not directly address the discrepancy in low-level control commands.
% == REDUCED ==
% Another line of work employs domain randomization~\citep{peng2018sim, andrychowicz2020learning, kumar2021rma, qi2023hand} to train under different robot dynamics so that the policy can adapt during deployment. Still, this assumes the availability of a simulation for the target task.
% == REDUCED ==
Domain randomization~\citep{peng2018sim, andrychowicz2020learning, kumar2021rma, qi2023hand} trains policies under varied robot dynamics to enable adaptation at deployment, but assumes that a simulation of the target task is available.

% === ORIGINAL ===
% \paragraph{Cross-embodiment policy learning}
% Several prior works learns policy by combining datasets from different embodiments~\citep{o2024openx, kim24openvla, brohan2023rt, zitkovich2023rt, bjorck2025gr00t, black2024pi0, intelligence2025pi}. 
% A few lines of work learn embodiment-specific action heads~\citep{bjorck2025gr00t, team2024octo, wen2025dexvla} or condition embodiment information~\citep{zheng2025x, chen2018hardware} information in the policy. Still, this requires fine-tuning to be deployed in a new embodiment and does not fundamentally address the dynamic differences between robots, but separates the effect by training per robot. 
% Meanwhile, our work directly uses a shareable action space across different embodiments and robot hardware without coupling dynamics dependency.
% Cosmos Policy~\citep{kim2026cosmos} and DreamZero~\citep{ye2026world} incorporate next frame prediction objective for policy learning, which can be shared across embodiments. However, they do not directly address the discrepancy in low-level control commands.
% Another line of works employ domain randomization~\citep{peng2018sim, andrychowicz2020learning, kumar2021rma, qi2023hand} to train under different robot dynamics so that policy can adapt during deployment. Still, this assumes the availability of a simulation for the target task.

\textbf{Learning from observations.~}
Prior works have attempted policy learning without relying on commands provided in datasets.
% GAIfO~\citep{torabi2018generative} uses online reinforcement learning to imitate observation sequences recorded in datasets. 
BCO~\citep{torabi2018behavioral} and SOIL~\citep{radosavovic2021state} learn a command-predicting policy by labeling commands from observation-only demonstrations using an inverse dynamics model of target robot.
Meanwhile, we train policy to directly output Cartesian state delta, which is universal across different robots.
SAIL~\citep{arachchige2025sail} and RACE~\citep{kim2025race} also learn policies that predict reached states in datasets instead of commands. Still, they do so for the purpose of speeding up policy execution and require high-stiffness gains~\citep{arachchige2025sail} or a path optimization algorithm~\citep{kim2025race} to execute the resulting policy.
Similar to our work, UMI~\citep{chi2024universal} and FastUMI~\citep{zhaxizhuoma2025fastumi} learn a policy to predict Cartesian state delta of a human hand-held gripper trajectory. However, they rely on engineering techniques such as latency measures to execute the resulting policy and focus on the transfer between human to robot only.

\section{Problem Formulation} 
\label{sec:pre}

We first describe robot policy learning via imitation learning and discuss common action space choices (Section~\ref{sec:action_spaces}). We then highlight the limitations of control commands as the action space when learning across different embodiments and hardware units (Section~\ref{sec:control_command_limit}).
% \vspace{-4pt}

\subsection{Preliminaries: Imitation Learning and Action Spaces} 
\label{sec:action_spaces}

We formulate robot control as a sequential decision-making problem in which a policy outputs an action given an observation. Formally, at timestep $t$, a policy $\pi$ receives an observation $o_t$ from the environment and outputs an action $a_t$. The observation $o_t = \{p_t, I_t, l_t\}$ consists of the robot's end-effector pose $p_t$ (or other proprioception input), a camera image $I_t$, and a language instruction $l_t$.
We adopt a behavior cloning~\citep{pomerleau1988alvinn} setting, in which the policy is trained from a precollected dataset of demonstrations. The dataset $\mathcal{D} = \{\tau^i\}_{i=1}^{N}$ consists of $N$ trajectories, where each trajectory $\tau^i = \{o_0, a_0, \dots, o_T, a_T\}$ is a sequence of observations and actions up to the terminal timestep $T$. 
% === REVISED ===
The policy is parameterized by a neural network with parameters $\theta$ and is trained to maximize the likelihood of the demonstrated actions as $\max_{\theta}\mathbb{E}_{\tau \sim \mathcal{D},\, (o_t, a_t) \sim \tau}\left[ \log \pi_\theta(a_t \mid o_t) \right]$.
% === ORIGINAL ===
% The policy is parameterized by a neural network with parameters $\theta$ and is trained to maximize the likelihood of the demonstrated actions:
% \begin{align*}
%     \max_{\theta} \; \mathbb{E}_{\tau \sim \mathcal{D},\, (o_t, a_t) \sim \tau}
%     \left[ \log \pi_\theta(a_t \mid o_t) \right].
% \end{align*}

% === REVISED ===
Prior work has explored a variety of action spaces for policy learning. Actions are typically defined along two axes: the \emph{modality} (joint or end-effector) and the \emph{temporal axis} (absolute or delta)~\citep{khazatsky2024droid, kim24openvla, intelligence2025pi, feng2026demystifying}. 
Combining these axes yields four commonly used action spaces. \emph{Joint absolute} commands a target joint position, while \emph{joint delta} commands a target joint displacement. \emph{Cartesian absolute} commands a target end-effector pose (Cartesian position and orientation), while \emph{Cartesian delta} commands a target pose displacement.\footnote{For manipulators, a gripper action is typically appended to each of these action spaces; we omit it from our notation for clarity.}
% === ORIGINAL ===
% Prior work has explored a variety of action spaces for policy learning. Actions are typically defined along two axes: the \emph{modality} (joint or end-effector) and the \emph{target type} (absolute or delta)~\citep{khazatsky2024droid, kim24openvla, intelligence2025pi, feng2026demystifying}. Combining these axes yields four commonly used action spaces. \emph{Joint absolute} commands a target joint position, while \emph{joint delta} commands a target joint displacement from the current joint position. \emph{Cartesian absolute} commands a target end-effector pose (Cartesian position and orientation), while \emph{Cartesian delta} commands a target pose displacement from the current end-effector pose.\footnote{For manipulators, a gripper action is typically appended to each of these action spaces; we omit it from our notation for clarity.}

\textbf{Control commands vs. achieved motion.~}
Beyond the modality and temporal axis, the action $a_t$ predicted by the policy is conventionally defined as the \emph{control command} sent to the robot's underlying controller (e.g., an operational-space controller~\citep{khatib1987unified}). However, the control command and the robot's actual motion are not the same: due to imperfect command tracking, a robot often moves less than what the command specifies, and the command thus should be larger than the desired motion to achieve it~\citep{bronars2026tune, arachchige2025sail, kim2025race}.
To make this concrete, consider the Cartesian delta action space. Let $u_t$ denote the Cartesian delta control command at timestep $t$, and let $p_t = (\mathbf{x}_t, \mathbf{r}_t)$ denote the end-effector pose, where $\mathbf{x}_t$ is its Cartesian position and $\mathbf{r}_t$ is its orientation in Euler angles. When the controller does not perfectly track the commanded motion, $\|u_t\| > \|\Delta p_t\|$ holds for the achieved motion $\Delta p_t = p_{t+1} - p_t$. 
This discrepancy is common across robots. 
% This discrepancy is common across robots and persists unless a stiff controller with carefully tuned gains is used~\citep{arachchige2025sail}. 
For instance, it occurs throughout DROID~\citep{khazatsky2024droid}, as we show in our experiments.
% \vspace{-4pt}

\subsection{Inconsistency of Control Commands across Robots}
\label{sec:control_command_limit}
Recent work has scaled robot learning by training policies on data from multiple embodiments~\citep{o2024openx, kim24openvla, team2024octo}, often using the Cartesian delta action space~\citep{kim24openvla, team2024octo} since it is less dependent on robot-specific kinematics and invariant to base-frame translation~\citep{feng2026demystifying, chen2024mirage}. In practice, this is typically realized by predicting Cartesian delta control commands $u_t$ that are fed to the underlying robot controller~\citep{kim24openvla, team2024octo}.

\begin{wrapfigure}{r}{0.5\textwidth}
    \centering
    \vspace{-10pt}
    \includegraphics[width=\linewidth]{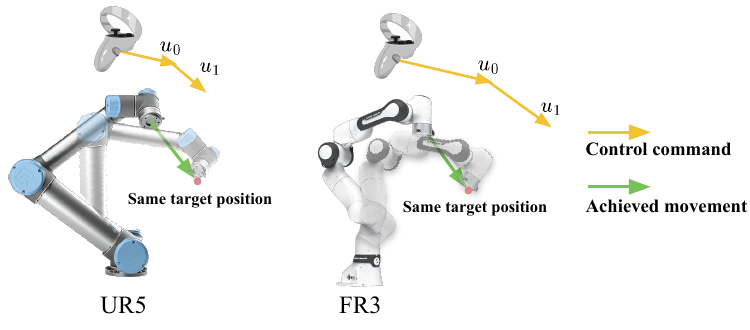}
    \caption{Different robots (e.g., UR5 vs. Franka Research 3) require different commands to perform the same movement.}
    \label{fig:problem_formulation}
    \vspace{-10pt}
\end{wrapfigure}

However, different robots require different control commands to achieve the same motion (Figure~\ref{fig:problem_formulation}). This holds not only across different embodiments, e.g., Franka Research 3 vs.~UR5, but even across different hardware units of the same embodiment, due to wear, manufacturing variability, and differences in low-level controllers or their gains~\citep{zheng2025universal, bronars2026tune, hao2021dynamic}. Consequently, control commands recorded in one robot's trajectory $\tau = \{o_0, u_0, o_1, u_1, \dots, o_T, u_T\}$ are not directly valid on other robots, harming both training across multiple robots and deployment on target robots.

Beyond inconsistency, Cartesian delta control commands are not even guaranteed to exist in many demonstration datasets. For example, leader-follower teleoperation systems~\citep{zhao2023aloha} often record control commands in the leader arm's joint space rather than in Cartesian space. Robots teleoperated via kinesthetic teaching~\citep{akgun2012kinesthetic, li2025train} produce no explicit control commands, since the robot is manipulated directly by a human.
% ; only the resulting motions are recorded. 
Similarly, human motion data and data collected with hand-held grippers (e.g., UMI~\citep{chi2024universal}) contain only recorded trajectories without explicit robot commands.
% === REVISED ===
In this work, we address these issues so that policies can effectively be trained and deployed on diverse robots.

\section{\Framename (\framename)}
% \section{Main Method}
\label{sec:method}
% === REVISED ===
In this section, we present \framename, a unified framework that consists of a Cartesian state delta policy and \algname that can handle dynamic variations across embodiments, across hardware units of the same robot embodiment, and within a single robot during operation.
% In this section, we present a unified framework that handles dynamic variations across embodiments, across hardware units of the same robot embodiment, and within a single robot during operation.
% === ORIGINAL ===
% In this section, we present a unified framework that handles dynamic variations in robotic control across embodiments, across hardware units of the same robot embodiment, and within a single robot during operation. Our framework is built around the cartesian state delta as a universal action interface, and consists of two components: (i) a cartesian state delta policy that predicts end-effector motion in this shared space (Section~\ref{sec:cart_pol}), and (ii) \algname that translates the predicted cartesian state delta into the control commands of a specific target robot.

\subsection{Cartesian State Delta Policy} \label{sec:cart_pol}
% Modern robot demonstration datasets are collected from heterogeneous sources: across different embodiments (e.g., Open-X~\citep{o2024openx}) and across many hardware units of the same robot model (e.g., DROID~\citep{khazatsky2024droid}). Since control commands and joint actions are not consistent across these robots, we instead train the policy to predict the actual robot motion recorded in each demonstration. Concretely, we predict the \emph{cartesian state delta} $\Delta p_t = p_{t+1} - p_t$, the displacement between current and next end-effector poses.\footnote{For the rotational component, deltas are computed by converting $\mathbf{r}_t$ and $\mathbf{r}_{t+1}$ to rotation matrices $R_t$ and $R_{t+1}$ to compute relative rotation as $\Delta R_t = R_{t+1}R_t^\top$. Then $\Delta R_t$ is converted back to Euler angles to obtain $\Delta \mathbf{r}_t$.}
% Unlike control commands or joint actions, the cartesian state delta is agnostic to robot dynamics since it expresses only end-effector motion. It can be obtained from any robot that provides end-effector cartesian poses, regardless of the control modality used during teleoperation. Moreover, it is invariant to robot kinematics and base-frame translation, and can therefore be shared across different embodiments and hardware units.
% This leads to the following behavior cloning objective:
Modern robot demonstration datasets are collected from heterogeneous sources: across different embodiments (e.g., Open-X~\citep{o2024openx}) and across many hardware units of the same robot embodiment (e.g., DROID~\citep{khazatsky2024droid}). 
Since control commands are not consistent across these robots, we instead train the policy to predict the actual robot motion recorded in each demonstration. 
Concretely, we predict the \emph{Cartesian state delta} $\Delta p_t = p_{t+1} - p_t$, the displacement between current and next end-effector poses.\footnote{For the rotational component, $\mathbf{r}_t$ and $\mathbf{r}_{t+1}$ are converted to rotation matrices $R_t$ and $R_{t+1}$ to compute the relative rotation $\Delta R_t = R_{t+1}R_t^\top$, which is then converted back to Euler angles to obtain $\Delta \mathbf{r}_t$.}
Since it expresses only end-effector motion, the Cartesian state delta is agnostic to robot dynamics. 
%Unlike control commands, the Cartesian state delta is agnostic to robot dynamics since it expresses only end-effector motion. 
It can be obtained from any robot that provides end-effector Cartesian poses, regardless of the control commands used during teleoperation. Moreover, it is invariant to robot kinematics and base-frame translation, and can therefore be shared across different embodiments and hardware units.
This leads to the following behavior cloning objective:
\begin{equation*}
\max_{\theta} \; 
\mathbb{E}_{\tau \sim \mathcal{D},\, (o_t, \Delta p_t) \sim \tau}
\left[
\log \pi_\theta(\Delta p_t \mid o_t)
\right],
\end{equation*}
where $o_t = \{p_t, I_t, l_t\}$ consists of the end-effector pose, camera image, and language instruction. We refer to a policy learned with this objective as a \emph{Cartesian state delta policy}.
% However, naively commanding the predicted Cartesian state delta from the policy may not produce the desired motion due to imperfect tracking, which we resolve in Section~\ref{sec:action_adapter} with \algname.
However, naively commanding the predicted Cartesian state delta from the policy does not always produce the desired motion on the target robot due to imperfect tracking (see Section~\ref{sec:action_spaces} for details). We resolve this in Section~\ref{sec:action_adapter} with \algname.
% However, executing a Cartesian state delta policy introduces a new challenge: naively commanding the predicted delta does not always produce the desired motion on the target robot. Robot controllers do not always track input commands perfectly, and the resulting motion can be substantially smaller than the predicted Cartesian state delta (see Section~\ref{sec:action_spaces} for details). We resolve this in Section~\ref{sec:action_adapter} with \algname.
% \input{algorithms/algorithm1}

\subsection{Action Adapter}
\label{sec:action_adapter}
% \algname translates the Cartesian state delta predicted by the policy into a control command for the target robot. Given a target robot, \algname learns a mapping from a Cartesian state delta $\Delta p_t$ to the control command $u_t$ that realizes it. We parameterize \algname as a linear model,
% \begin{equation*}
% u_t = W_0 \Delta p_t + b_0,
% \end{equation*}
% === NEW ===
\algname translates the Cartesian state delta predicted by the policy into a control command for a target robot. Given a target robot, \algname learns a mapping from a Cartesian state delta $\Delta p$ to the control command $u$ that realizes it. \algname is parameterized as a linear model,
% === ORIGINAL ===
% \algname translates the Cartesian state delta predicted by the policy into a control command for the target robot. Given a target robot, \algname learns a mapping from a Cartesian state delta $\Delta p$ to the control command $u$ that realizes it. We parameterize \algname as a linear model,
\begin{equation*}
\hat{u} = W_0 \Delta p + b_0,
\end{equation*}
and fit its parameters by minimizing a least-squares objective:
\begin{equation}
\min_{W_0, b_0} \;
\sum_{(\Delta p, u) \in \mathcal{D}_{\text{cal}}}
\left\| W_0 \Delta p + b_0 - u \right\|_2^2,
\label{eq:adapter_linear_regression}
\end{equation}
where $\mathcal{D}_{\text{cal}}$ denotes the calibration trajectories used to train \algname. 
% === REVISED ===
To collect $\mathcal{D}_{\text{cal}}$, we roll out a random scripted policy $\pi_\text{rand}$ and gather $M$ trajectories of length $K$. 
% === ORIGINAL ===
% To collect $\mathcal{D}_{\text{cal}}$, we roll out a random scripted policy $\pi_\text{rand}$ and gather $M$ trajectories of length $K$, i.e., $\mathcal{D}_\text{cal} = \{\tau^i_\text{rand}\}_{i=1}^{M}$ where each $\tau^i_\text{rand} = \{p_0, u_0, \dots, p_K, u_K\}$. 
This calibration step typically takes less than one minute on a Franka Research 3 robot with $M=10$ and $K=50$, and \algname is fit in negligible time via linear regression.
% (see Appendix~\ref{app:adapter_ablation} for supporting results)
However, as the policy rolls out, changes in robot configuration can make the learned parameters $W_0$ and $b_0$ inaccurate. For example, when the robot approaches a singularity or holds a heavy object, the same control command $u_t$ may produce a smaller actual movement. 

To handle this, we continuously update \algname during deployment using the least mean squares (LMS) algorithm~\citep{haykin2003least}. Let $\pi_\theta$ be the learned Cartesian state delta policy. At rollout timestep $t$, the policy outputs $\Delta p_t^\text{target} \sim \pi_\theta(\cdot \mid o_t)$, which \algname converts into a control command $\hat{u}_t = W_t \Delta p_t^\text{target} + b_t$. After executing $\hat{u}_t$, we observe the actual robot motion $\Delta p_t^\text{obs} = p_{t+1} - p_t$ and define the prediction error as $e_t = W_t \Delta p_t^\text{obs} + b_t - \hat{u}_t$. Note that the error is defined using $\Delta p_t^\text{obs}$ rather than $\Delta p_t^\text{target}$, since no ground-truth control command is available for $\Delta p_t^\text{target}$. The adapter is then updated via LMS:
\begin{equation*}
W_{t+1} = W_t - \mu e_t (\Delta p_t^\text{obs})^\top,
\qquad
b_{t+1} = b_t - \mu e_t,
\end{equation*}
where $\mu$ is the learning rate. This update is equivalent to one step of online gradient descent on the squared error $\|e_t\|_2^2$. 
Algorithm~\ref{alg:action_adapter} in Appendix~\ref{app:pseudo_code}  summarizes the full procedure for executing the Cartesian state delta policy with \algname.

\section{Experiments} \label{sec:experiments}

We design our experiments to investigate the following:
\begin{enumerate}[label=(\roman*), leftmargin=*, itemsep=3pt, topsep=2pt, parsep=0pt]
    \item Does \framename improve performance over a policy predicting control commands in cross-embodiment learning? (Section~\ref{sec:cross_emb})
    \item Does \framename improve performance when learning across multiple hardware units of the same embodiment? (Section~\ref{sec:cross_hardware})
    \item Does \framename remain robust under a dynamics shift at deployment? (Section~\ref{sec:dynamics_change})
\end{enumerate}

% === REVISED ====
We conduct all experiments on a real Franka Research 3 (FR3) robot using the DROID~\citep{khazatsky2024droid} platform, a widely used FR3 robot setup~\citep{kim24openvla, black2024pi0, ye2026world, atreya2025roboarena}. We mainly compare \framename against a policy that predicts control commands, which we refer to as the \emph{control command policy}.
We adopt $\pi_{0.5}$~\citep{intelligence2025pi}, a state-of-the-art vision-language-action model, as our main policy and fine-tune it for 20k steps with a batch size of 64 unless otherwise specified. To initialize the weights $W_0$ and $b_0$ of \algname, we collect $M=10$ random trajectories of length $K=50$ and perform linear regression. The learning rate of \algname is set to $\mu = 0.2$ across all experiments. We refer readers to Appendix~\ref{app:exp_details} for further details in experiments.
% We conduct all experiments on a real Franka Research 3 robot (abbreviated as Franka) using the DROID~\citep{khazatsky2024droid} platform, a widely used Franka robot setup~\citep{kim24openvla, black2024pi0, ye2026world, atreya2025roboarena}. We mainly compare the Cartesian state delta policy against a policy that predicts control commands, which we refer to as the \emph{control command policy}.

We use the robot controller from DROID~\citep{khazatsky2024droid} with the Cartesian delta control command modality, which commands the desired end-effector pose displacement.\footnote{Note that this is different from the ``Cartesian velocity'' command in DROID, which is a normalized version of the ``Cartesian delta'' command in $[-1, 1]$.} \algname therefore converts the predicted Cartesian state delta into a Cartesian delta control command. The success rate is evaluated over 50 rollouts, and \algname is reset to the initial $W_0$ and $b_0$ fitted from $\mathcal{D}_\text{cal}$ before each rollout to ensure independence between rollouts.

\begin{figure}[ht]
    \centering
    % \vspace{-5pt}
    % \includegraphics[width=0.7\textwidth]{figures/figure_cross_embodiment_environment.pdf}
    \includegraphics[width=0.88\textwidth]{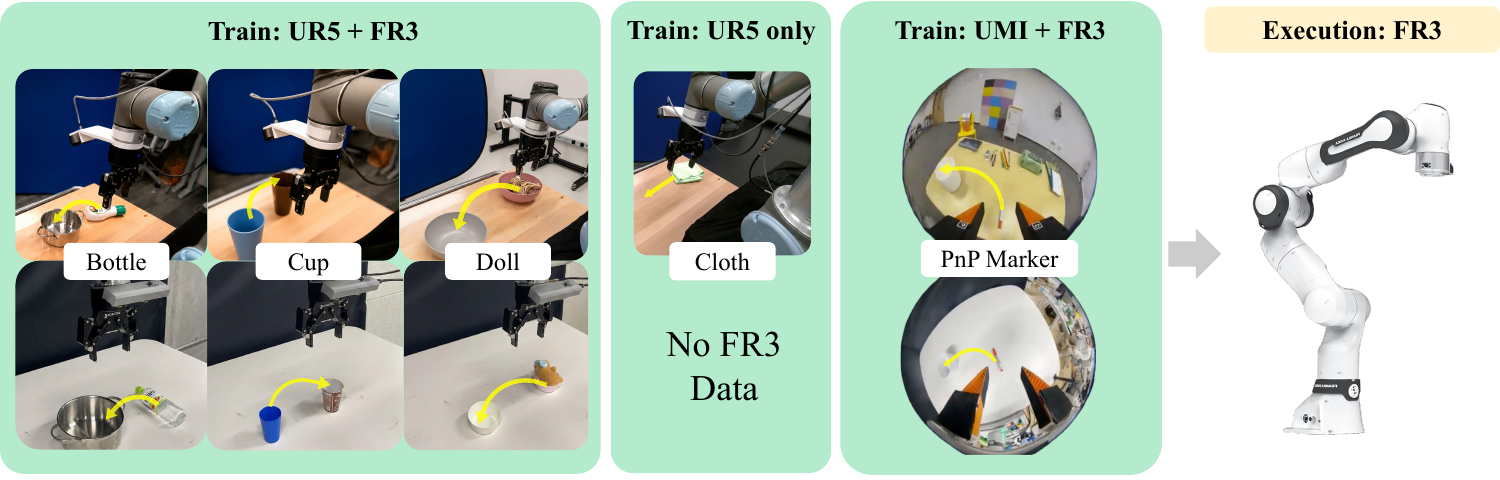}
    \caption{We study co-training between UR5 and FR3 robot for 3 different tasks, and zero-shot transfer from UR5 to FR3 for the Cloth task. We also study the transfer between data collected by a human hand-held gripper (UMI) and the FR3 robot in the pick-and-place (PnP) marker task.}
    \label{fig:cross_embodiment_env}
    \vspace{-8pt}
\end{figure}

\subsection{Does \framename Improve Cross-Embodiment Learning?}
% \subsection{Does the Cartesian State Delta Policy Improve Cross-Embodiment Learning?}
\label{sec:cross_emb}
We evaluate whether \framename more effectively leverages datasets from different embodiments than the control command policy.
% We evaluate whether the Cartesian state delta policy, executed with \algname, more effectively leverages datasets from different embodiments than the control command policy.

\textbf{Transfer from UR5 to FR3.~}
We use 250 demonstrations from the UR5 robot in the Berkeley UR5 Demonstration dataset~\citep{BerkeleyUR5} to train policies for the FR3 robot on four tasks: \emph{Bottle}, \emph{Cup}, \emph{Doll}, and \emph{Cloth}. As shown in Figure~\ref{fig:cross_embodiment_env}, the robot places a bottle into the pot (Bottle), stacks a blue cup on top of the brown cup (Cup), places a doll from the red bowl into the white bowl (Doll), and sweeps the cloth to the left side of the table (Cloth). 
% For the Bottle, Cup, and Doll tasks, we collect 20 FR3 demonstrations and co-train with the 250 UR5 demonstrations with balanced batch. For the Cloth task, we use only UR5 data and evaluate zero-shot transfer to the FR3.
For the Bottle, Cup, and Doll tasks, we collect 20 FR3 demonstrations and co-train with the 250 UR5 demonstrations, balancing the datasets so FR3 and UR5 are sampled equally per batch. For the Cloth task, we use only UR5 data and evaluate zero-shot transfer to the FR3.
\begin{figure}[H]
    \centering
    \begin{subfigure}[t]{0.47\textwidth}
        \centering
        \includegraphics[width=\linewidth]{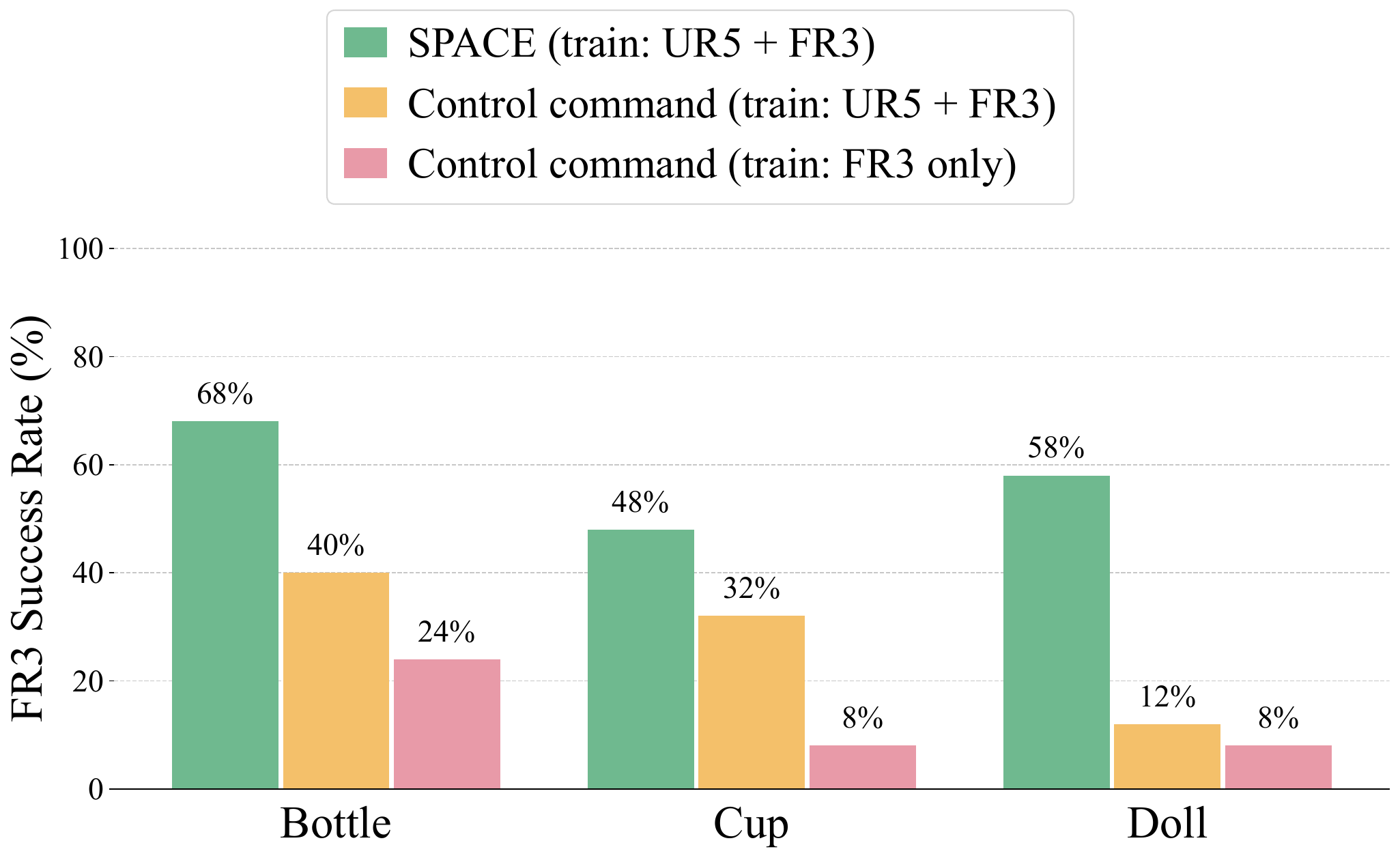}
        \caption{UR5 + FR3 co-training}
        \label{fig:ur5_cotraining}
    \end{subfigure}
    \hspace{0.01\textwidth}
    % \hfill
    \begin{subfigure}[t]{0.23\textwidth}
        \centering
        \includegraphics[width=\linewidth]{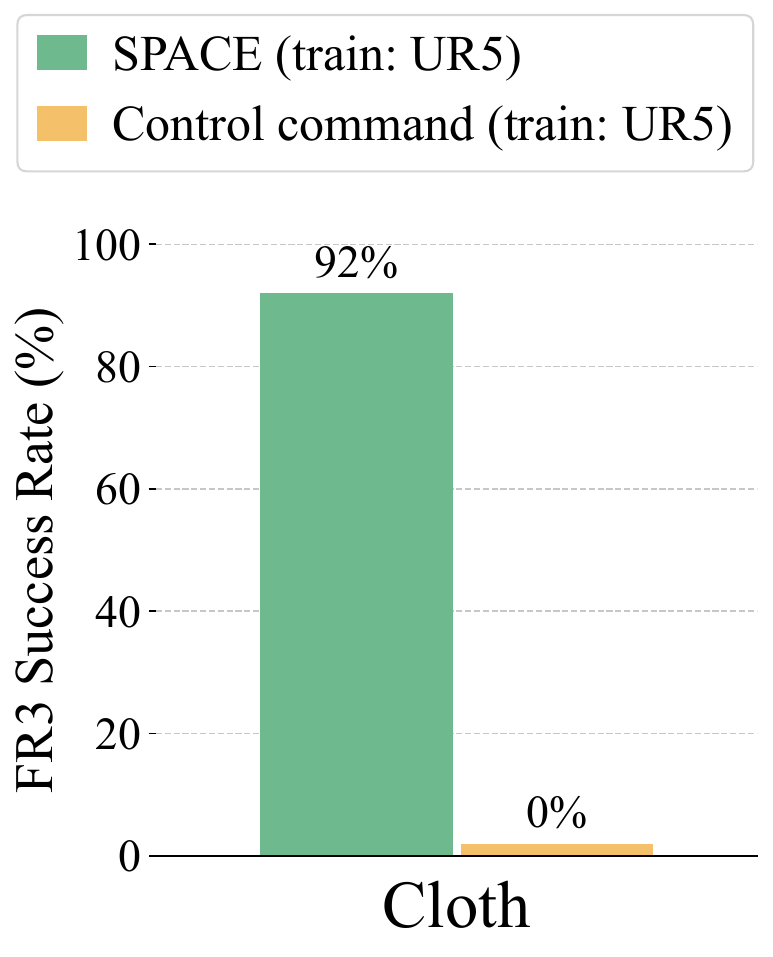}
        \caption{UR5 only training}
        \label{fig:ur5_zero_shot}
    \end{subfigure}
    \hspace{0.01\textwidth}
    % \hfill
    \begin{subfigure}[t]{0.25\textwidth}
        \centering
        \includegraphics[width=\linewidth]{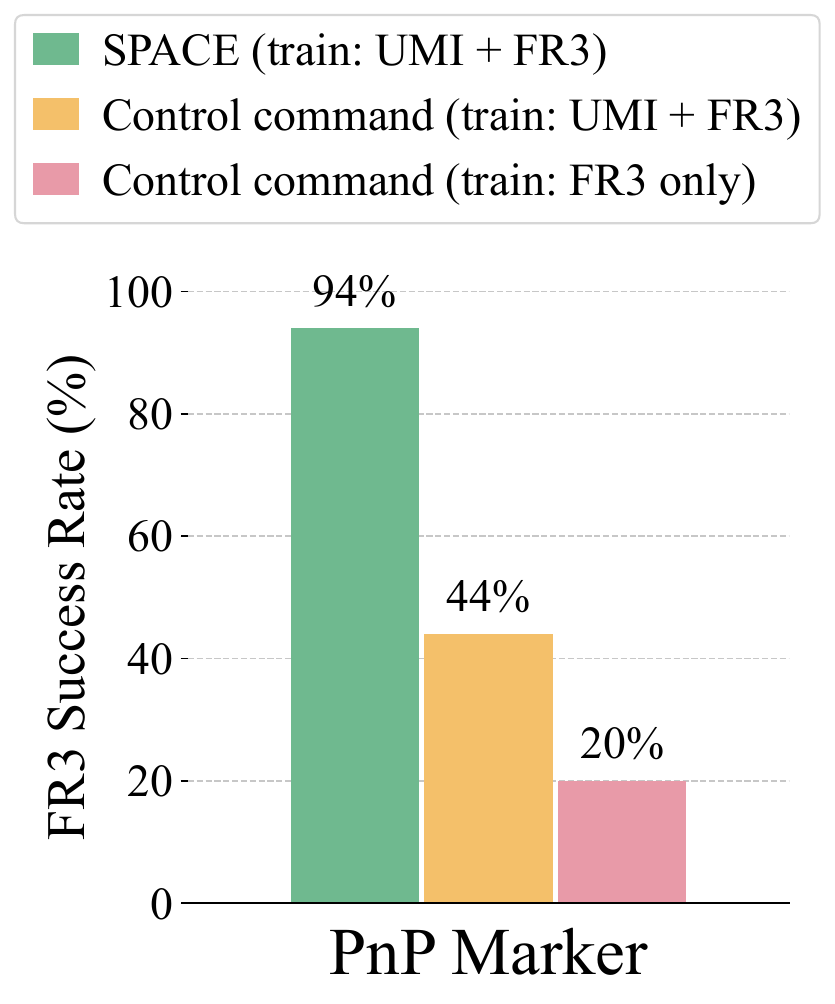}
        \caption{UMI + FR3 co-training}
        \label{fig:human_cotraining}
    \end{subfigure}
    \caption{Success rate for cross-embodiment experiments. We study (a) co-training on FR3 and UR5 data, (b) zero-shot transfer from UR5 to FR3, and (c) co-training on UMI and FR3 data.}
    \label{fig:cross_embodiment_result}
    \vspace{-15pt}
\end{figure}

In the co-training setting (Figure~\ref{fig:ur5_cotraining}), \framename achieves the highest success rate across all three tasks, outperforming both co-training with the control command policy and training on FR3 data alone, with an average improvement of 30\% over the control command policy. Moreover, when trained solely on the UR5 dataset, \framename enables zero-shot transfer with a 92\% success rate on the Cloth task, whereas training with UR5 control commands leads to significant under-reaching in the FR3 robot, resulting in a 0\% success rate (Figure~\ref{fig:ur5_zero_shot}).
% To verify whether our framework is responsible for these gains, we perform a replay analysis on the FR3 robot. 
In addition, replaying UR5 trajectories using recorded Cartesian state deltas with \algname achieves the lowest tracking errors, whereas replaying UR5 control commands leads to large deviations in FR3 (Appendix~\ref{app:tracking_test}). 
This confirms that \framename effectively bridges the control command discrepancy between UR5 and FR3.
In addition, directly executing the Cartesian state delta policy without \algname achieves a 0\% success rate across all tasks due to under-reaching, highlighting the necessity of \algname.

\textbf{Transfer from human hand-held gripper data.~}
We further test whether \framename is also effective for utilizing hand-held gripper data, specifically from the FastUMI dataset~\citep{zhaxizhuoma2025fastumi} (which we refer to as ``UMI data'' for simplicity).
Since UMI data lacks control command, we naturally adopt the Cartesian state delta as an action space.
% We study co-training with UMI and robot data to enable execution in deployment settings beyond those captured by UMI data alone.
We study co-training with UMI and robot data to enable execution in different domains beyond UMI data collection.
We use the ``PnP marker'' task, which comprises 550 UMI trajectories in which a marker is grasped and placed into a bowl, and co-train with 10 FR3 robot trajectories collected for the same task.
As shown in Figure~\ref{fig:human_cotraining}, \framename (i.e., co-training with Cartesian state delta) outperforms co-training with control command from robot data by 50\%.
This confirms that using the Cartesian state delta as a robot data action leads to better transfer with UMI data by matching the data action space.

\subsection{Does \framename Improve Cross-Hardware Learning?}
\label{sec:cross_hardware}
In this section, we test whether \framename improves performance when training and deploying across different hardware units of the same embodiment.

\begin{wrapfigure}{r}{0.45\linewidth}
\centering
\includegraphics[width=0.95\linewidth]{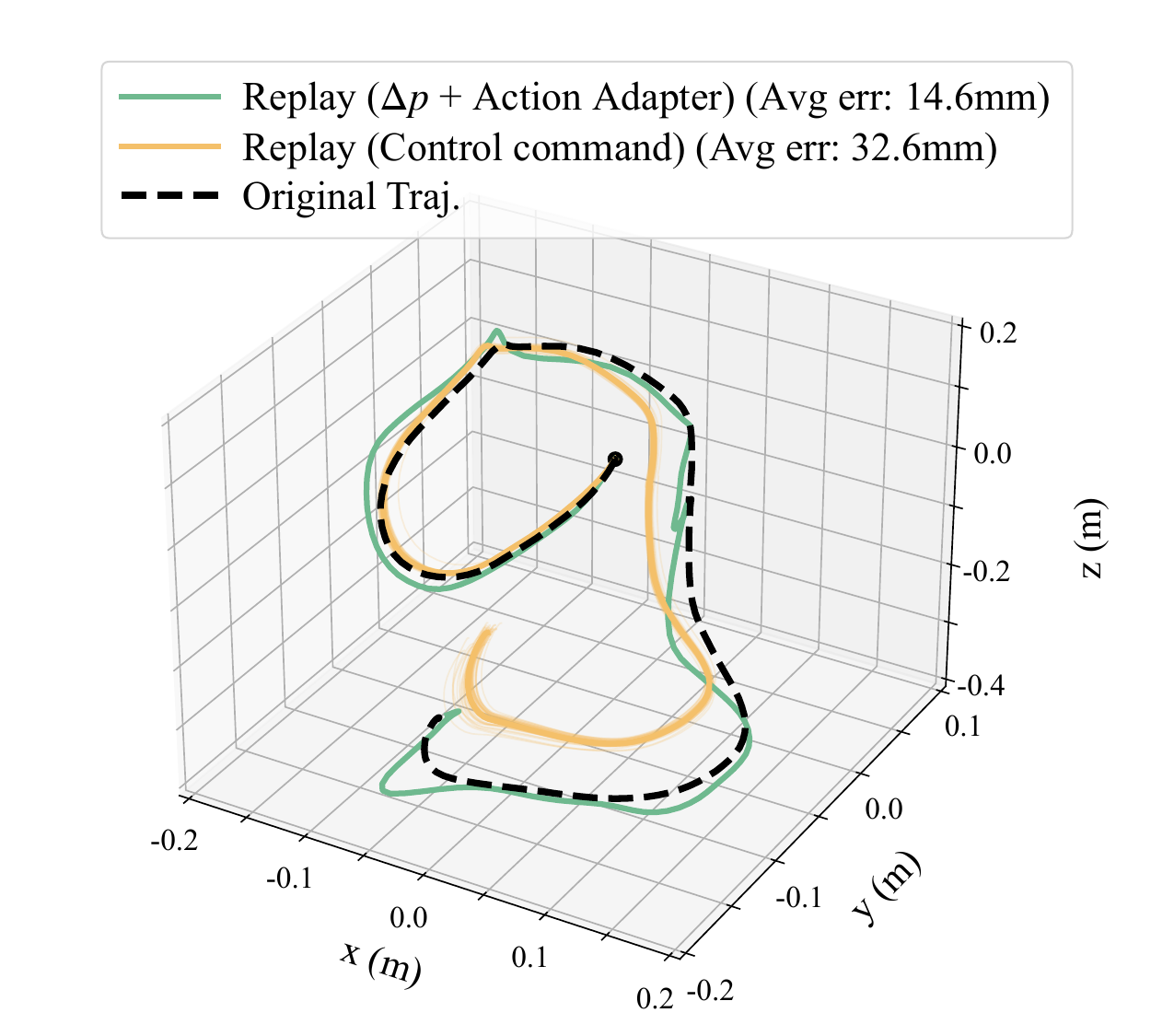}
\caption{FR3 trajectory replayed on a different hardware from the collection time. Errors are averaged over timesteps and trajectories, and $\Delta p$ denotes Cartesian state delta.}
% We collect the trajectory in Robot 1 and replay it in Robot 2.
\label{fig:replay_error}
\vspace{-10pt}
\end{wrapfigure}

\paragraph{Hardware-induced dynamics variation.}
Even though the embodiment (i.e., robot model) and controller implementation are the same, we hypothesize that different hardware units exhibit discrepancies in dynamics due to wear, manufacturing variability, and subtle different in setup (e.g., cable tension). To demonstrate this, we collect a single trajectory of 180 timesteps from FR3 Robot 1 and replay it on FR3 Robot 2 from the same initial pose 50 times. For each replay, we either send the original control commands or execute the Cartesian state deltas ($\Delta p$) through \algname. Figure~\ref{fig:replay_error} shows the resulting trajectories averaged over the 50 replays. Replaying with control commands produces a position tracking error of 32.6\,mm on Robot 2, much larger than the 6.3\,mm error on Robot 1. This shows that control commands recorded on Robot 1 produce less accurate motion when replayed on Robot 2, due to differences in dynamics. In contrast, replaying with Cartesian state deltas through \algname on Robot 2 yields a substantially smaller error, demonstrating our framework's ability to adapt to the new hardware.
% \begin{figure}[H]
%     \centering
%     \includegraphics[width=0.7\linewidth]{figures/figure_replay_error.pdf}
%     \caption{
%         % \footnotesize
%         Franka trajectory replayed on different hardware from collection time.
%         Errors are averaged over timesteps and trajectories.
%     }
%     \label{fig:replay_error}
%     \vspace{-10pt}
% \end{figure}
% We test whether \framename improves performance when training and deploying across different hardware units of the same embodiment.
% Even though the embodiment and controller implementation are the same, we hypothesize that different hardware units exhibit discrepancies in dynamics due to wear, manufacturing variability, and subtle differences in setup (e.g., cable tension).

\begin{figure}[ht]
    \vspace{-6pt}
    \centering
    \begin{minipage}{0.42\textwidth}
        \centering \includegraphics[width=\textwidth]{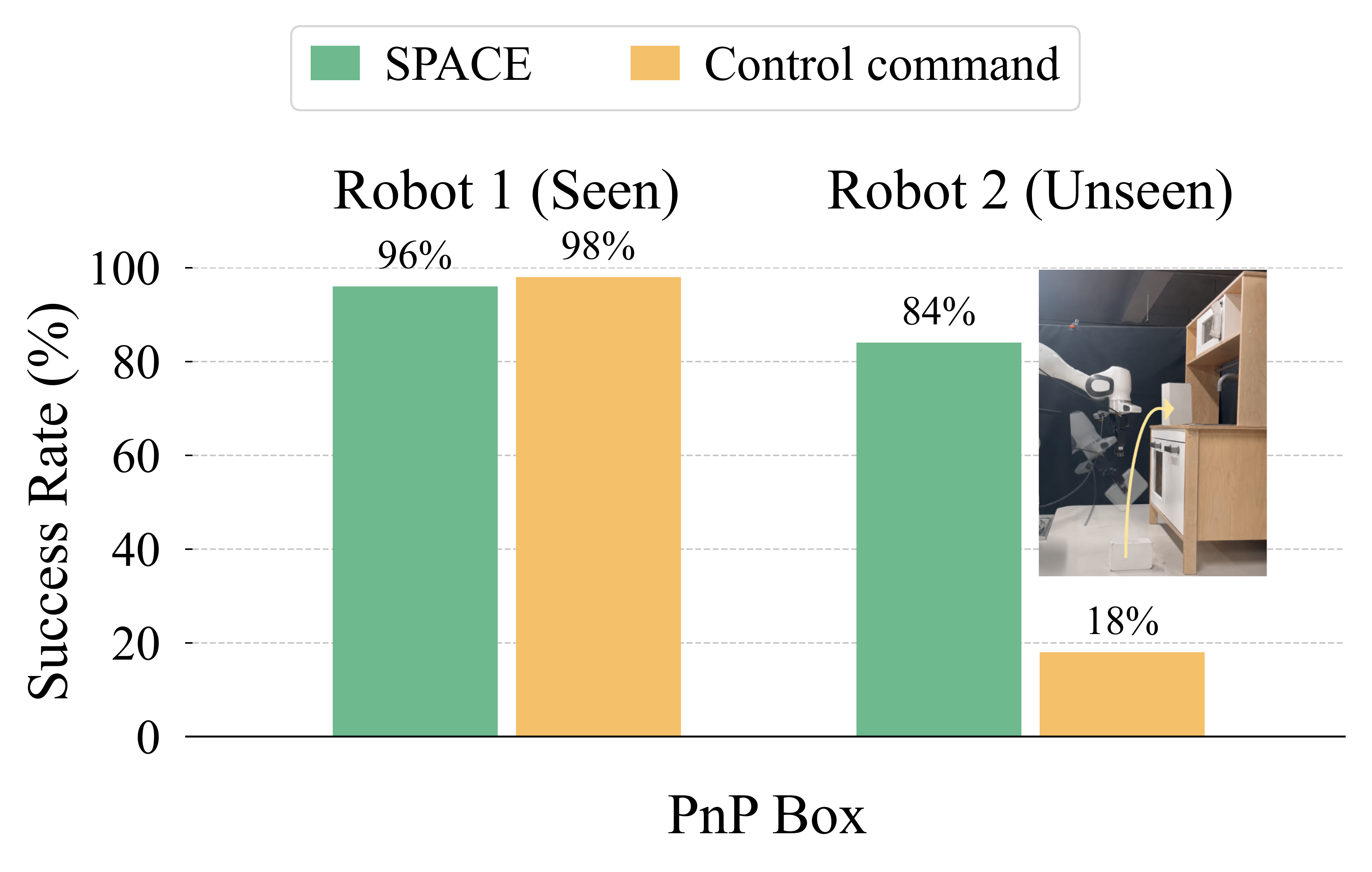}
        \captionof{figure}{Success rate on new hardware. Data is collected on Robot 1; the policy is evaluated on Robot 1 and Robot 2.}
        % \textbf{Droid Success Rate.}}
        \label{fig:hardware_success_rate}
    \end{minipage}
    \hfill
    \begin{minipage}{0.54\textwidth}
        \centering
        \includegraphics[width=\textwidth]{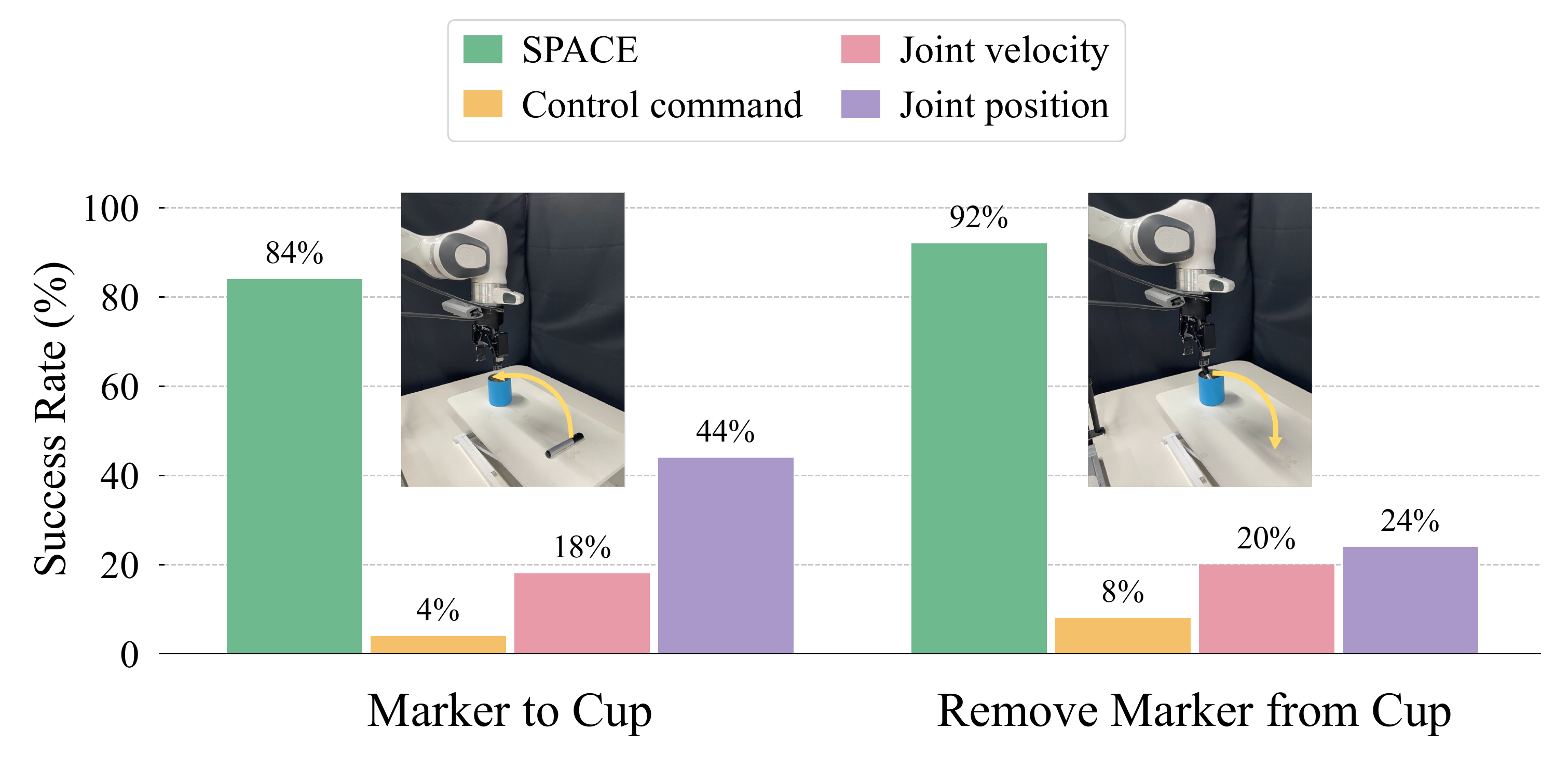}
        \caption{
        % \footnotesize
        % \textbf{DROID success rate.} Cartesian state delta policy outperforms control command policy when training from a subset of DROID dataset.}
        % $\mathbf{\pi_{0.5}}$ model
        Success rate in DROID. We compare \framename against policies trained using different action spaces available in DROID.}
        % Cartesian state delta policy outperforms control command policy when training from a subset of DROID dataset. }
        \label{fig:droid_success_rate}
        % \textbf{Different Hardware Success Rate.}}
    \end{minipage}
\vspace{-10pt}
\end{figure}

\textbf{Transferring a policy across hardware units.~}
To evaluate the impact of these dynamics differences on policy performance, we compare \framename and the control command policy, trained using 50 demonstrations of the PnP Box task where the robot moves a box from the floor onto the desk. The resulting policies are evaluated on both the original robot used for data collection (Seen) and a new hardware unit (Unseen). 
As shown in Figure~\ref{fig:hardware_success_rate}, the control command policy's success rate drops from 98\% on the original robot to 18\% on the new hardware unit. This confirms that control commands are coupled to the data collection robot's dynamics and fail to generalize across hardware. In contrast, \framename remains relatively robust, achieving an 84\% success rate on the new hardware unit by learning a Cartesian state delta that is agnostic to data collection dynamics.
% In Appendix~\ref{app:tracking_test}, we also show that replaying a trajectory using control command in a different robot from data collection increases tracking error, while replaying Cartesian state delta with \algname does not.

\textbf{Learning from multi-hardware data.~}
While the above results show that \framename enables transfer from one hardware unit to another, an equally important scenario is training a policy on data collected from many hardware units simultaneously.
To test this, we use the DROID dataset~\citep{khazatsky2024droid}, a large-scale open-source dataset of FR3 robot demonstrations collected across multiple labs and hardware units. Since DROID aggregates data from many hardware units, the control commands reflect a mixture of dynamics, providing a noisier supervision signal for the policy.
To reduce training cost, we filter DROID trajectories whose instructions contain ``marker", yielding 6614 trajectories with 1.4M samples, on which we fine-tune $\pi_{0.5}$ for 5 epochs with a batch size of 64.
The learned policy is evaluated on two tasks, ``Marker to Cup'' and ``Remove Marker from Cup'', in which the robot moves a marker into a cup on the tray and moves a marker inside the cup onto the tray, respectively. 
As shown in Figure~\ref{fig:droid_success_rate}, \framename significantly outperforms the control command policy, with gaps of 80\% on Marker to Cup and 84\% on Remove Marker from Cup.\footnote{In the DROID experiment, the control command policy uses the Cartesian velocity command, since the original DROID dataset does not provide Cartesian delta commands.} It also outperforms policies using other control command modalities such as joint velocity and joint position. 
This confirms \framename addresses subtle dynamics differences across hardware units that affect policy performance.

\subsection{Does \framename Work under Dynamics Shift from Training Time?}
% \subsection{Does the Cartesian State Delta Policy with \algname Work under Dynamics Shift?}
\label{sec:dynamics_change}
We also test whether \framename can handle dynamics variations within a single robot during operation.
% We also test whether \framename can handle dynamics variations within a single robot during operation, beyond those seen during data collection. 
These include changes in control frequency relative to training, as well as environmental variations such as object weight and controller gains. For example, increasing the control frequency above that used during data collection sends control commands to the robot more frequently, causing it to execute faster.
% We also test whether the Cartesian state delta policy can handle dynamics variations within a single robot during operation, beyond those seen during data collection. These include changes in control frequency relative to training, as well as environmental variations such as object weight and controller gains. For example, increasing the control frequency above that used during data collection sends control commands to the robot more frequently, causing it to execute faster.
% \input{figures/fig_hz_change}
% \input{figures/fig_adapter_visualization}

\begin{figure}[H]
    \vspace{-4pt}
    \centering
    \begin{minipage}{0.47\textwidth}
        \centering \includegraphics[width=\textwidth]{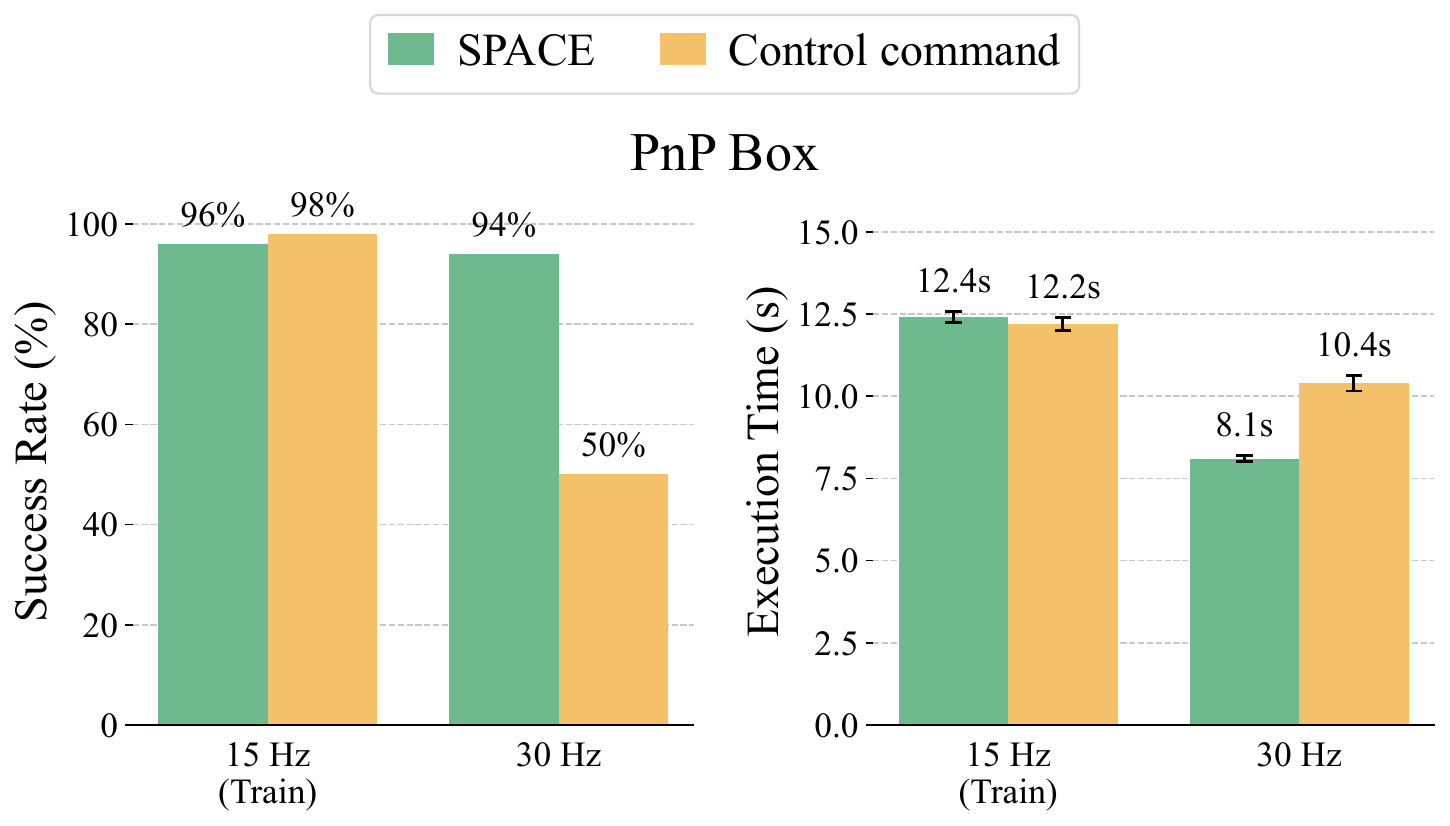}
        \captionof{figure}{
        % \footnotesize
        % $\mathbf{\pi_{0.5}}$ model
        Success rate and time taken until task completion in different execution Hz.}
        \label{fig:hz_variation}
    \end{minipage}
    \hspace{0.02\textwidth}
    % \hfill
    \begin{minipage}{0.45\textwidth}
        \centering
        \includegraphics[width=\textwidth]{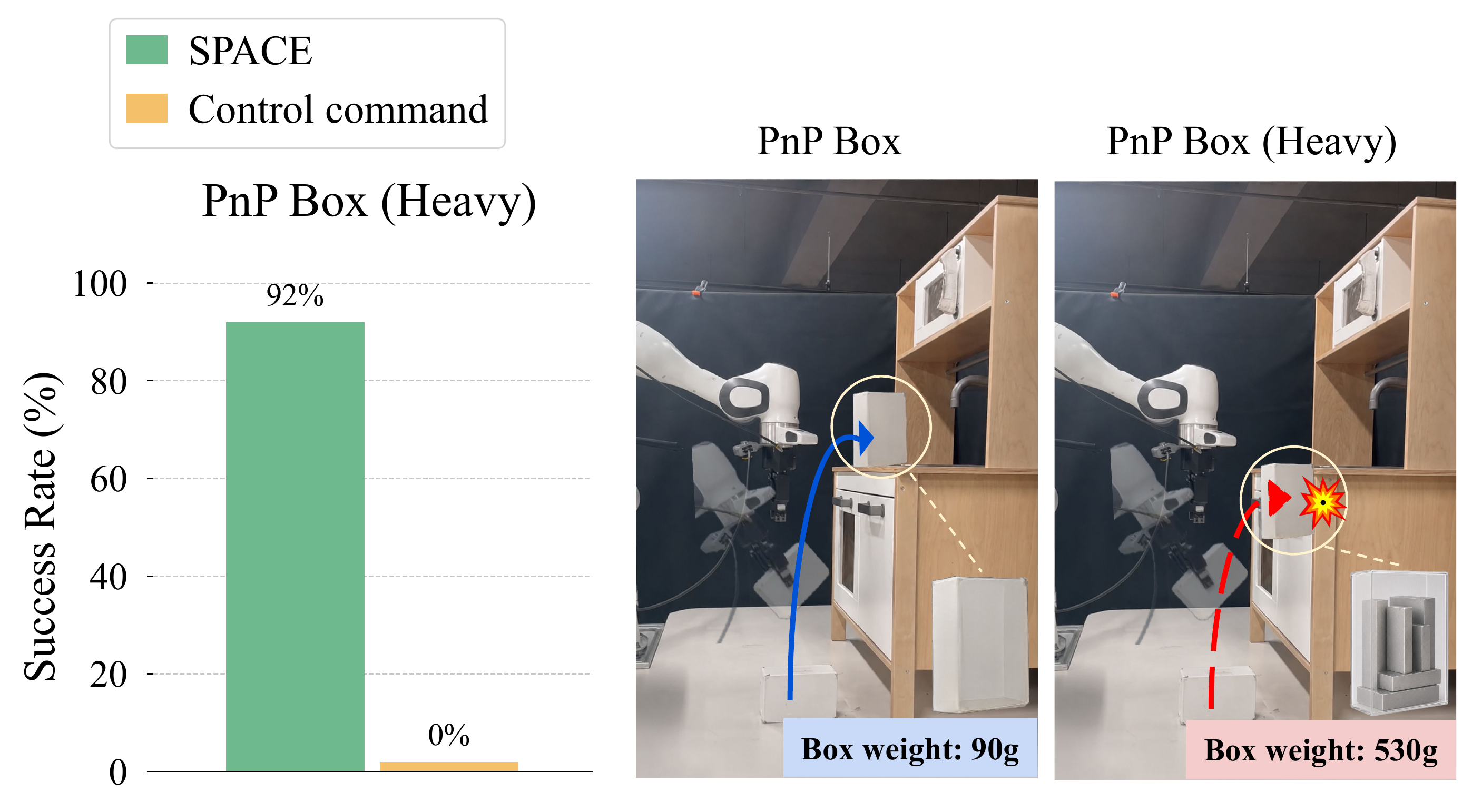}
        \caption{
        % \footnotesize
        % $\mathbf{\pi_{0.5}}$ model
        Success rate after increasing the object weight.}
        \label{fig:heavy_box}
    \end{minipage}
\vspace{-15pt}
\end{figure}

Specifically, we take the $\pi_{0.5}$ model trained on the PnP Box task in Section~\ref{sec:cross_hardware} and increase the execution frequency from 15Hz (used during data collection) to 30Hz. As shown in Figure~\ref{fig:hz_variation}, \framename allows execution at this higher frequency: the task completion time is significantly reduced from 12.4s to 8.1s, while maintaining task performance. In contrast, increasing the execution frequency for the control command policy causes a 48\% drop in success rate. This is because the control commands are tied to the control frequency used during data collection; executing the same commands at a higher frequency causes the robot to under-reach. Unlike prior work that accelerates policy execution by increasing the control frequency~\citep{arachchige2025sail, kim2025race}, \framename achieves this without controller gain tuning~\citep{arachchige2025sail} or path optimization algorithm~\citep{kim2025race}.
% Specifically, we take the $\pi_{0.5}$ model trained on the PnP Box task in Section~\ref{sec:cross_hardware} and increase the execution frequency from 15Hz (used during data collection) to 30Hz. As shown in Figure~\ref{fig:hz_variation}, the Cartesian state delta policy with \algname allows execution at this higher frequency: the task completion time is significantly reduced from \textcolor{blue}{12.4s} to \textcolor{blue}{8.1s}, while maintaining task performance. In contrast, increasing the execution frequency for the control command policy causes a 48\% drop in success rate. This is because the control commands are tied to the control frequency used during data collection; executing the same commands at a higher frequency causes the robot to under-reach. Unlike prior work that accelerates policy execution by increasing the control frequency~\citep{arachchige2025sail, kim2025race}, our method achieves this without controller gain tuning~\citep{arachchige2025sail} or path optimization~\citep{kim2025race}.
% \begin{figure}[H]
%     \centering
%     % \vspace{-.5in}
%     \includegraphics[width=\textwidth]{figures/figure_success_rate_hz.pdf}
%     \caption{
%         \footnotesize
%         \textbf{Success rate for Franka robot in different execution hz.}
%     }
%     \label{fig:hz_variation}
%     % Heavy weight via separate figure here.
%     % \label{fig:ur5_success_rate}
%     \vspace{-.15in}
% \end{figure}
\begin{wrapfigure}{r}{0.4\linewidth}
    \centering
    % \vspace{-2mm}
    \includegraphics[width=\linewidth]{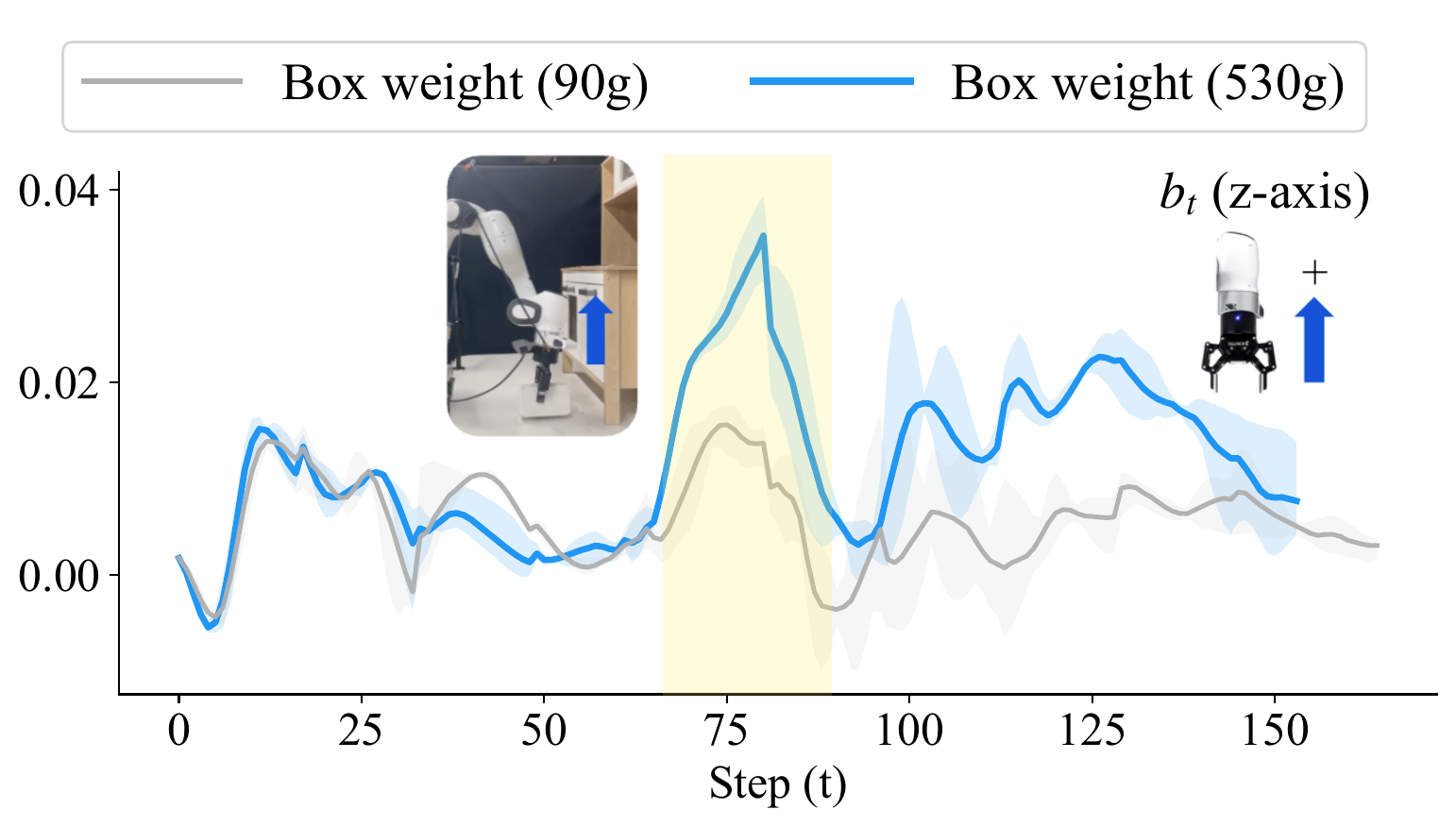}
    \caption{
        % \footnotesize
        \algname bias ($b_t$) z-axis visualization in PnP Box task.
        Heavy box weight leads to an increase in $z$-axis value after grasping (highlighted by yellow). The value is averaged over three rollouts. 
    }
    \label{fig:adapter_visualization}
    \vspace{-4mm}
\end{wrapfigure}

We also show that the \framename can handle dynamics changes driven by environmental factors such as object weight. Specifically, we use the same PnP Box task and add metal pieces to the box, increasing its weight from 90g to 530g. As shown in Figure~\ref{fig:heavy_box}, this drops the control command policy's success rate to 0\%; qualitatively, the robot grasps the box but fails to lift it. In contrast, \framename adapts its control commands online to compensate for the heavier box, achieving a 92\% success rate, closely matching that of the empty-box case.
To examine this adaptation, we visualize the $z$-component of the bias $b_t$ in \algname ($W_t \Delta p + b_t$) across timestep $t$ during online rollout, which captures the upward correction applied against gravity.
As shown in Figure~\ref{fig:adapter_visualization}, this component rises above $0.03$ after the heavy box is grasped (highlighted in yellow), compared to under $0.02$ for the empty box. This indicates that the online update actively compensates for the added weight.
% A complementary result in Appendix~\ref{app:controller_gains} shows that \framename also remains robust to controller gains that differ from training.

We also test whether \framename remains robust when executed with different controller gains from training time.
Using the same PnP Box task, we vary proportional gains ($K_p$) in DROID controller~\citep{khazatsky2024droid} by $0.5\times$ and $1.5\times$ from the training time.
Table~\ref{tab:franka_gain_change} displays policy success rates under those changed gains measured over 20 rollouts.
% We evaluate policy success rates using 20 rollouts under those changed gains, except for $K_p$ (1$\times$), for which we use 50 rollouts.

% \begin{table}[H]
% \centering
% \small
% \setlength{\tabcolsep}{4pt}
% \renewcommand{\arraystretch}{1.1}
% \begin{tabular}{lccc}
% \toprule
% \textbf{PnP Box}
% & \textbf{Kp (1$\times$)} & \textbf{Kp (0.5$\times$)} & \textbf{Kp (1.5$\times$)} \\
% \midrule
% % Cartesian state delta           & 0\% & 0\% & 0\% \\
% Control command                 & 98\% & 25\% & 0\% \\
% SPACE & 96\% & 95\% & 100\% \\
% \bottomrule
% \end{tabular}
% \vspace{6pt}
% \caption{Success rates on the PnP Box task under controller stiffness gains multiplied by different factors from training. 20 rollouts are used except for Kp (1$\times$) using 50 rollouts to measure the success rate.}
% \vspace{-10pt}
% \label{tab:franka_gain_change}
% \end{table}
% \begin{wraptable}{r}{0.54\textwidth}
%     \centering
%     \small
%     \setlength{\tabcolsep}{3pt}
%     \renewcommand{\arraystretch}{1.1}
%     \begin{tabular}{lccc}
%         \toprule
%         \textbf{PnP Box}
%         & \textbf{$K_p$ (1$\times$)}
%         & \textbf{$K_p$ (0.5$\times$)}
%         & \textbf{$K_p$ (1.5$\times$)} \\
%         \midrule
%         SPACE           & 96\% & 95\% & 100\% \\
%         Control command & 98\% & 25\% & 0\% \\
%         \bottomrule
%     \end{tabular}
%     \caption{Success rates on the PnP Box task under controller stiffness gains multiplied by different factors from training.}
%     \label{tab:franka_gain_change}
%     \vspace{-10pt}
% \end{wraptable}
\begin{wraptable}{r}{0.54\textwidth}
    \centering
    \small
    \setlength{\tabcolsep}{3pt}
    \renewcommand{\arraystretch}{1.1}
    \begin{tabular}{lccc}
        \toprule
        \textbf{PnP Box}
        & \textbf{$K_p$ (Train)}
        & \textbf{$K_p$ (0.5$\times$)}
        & \textbf{$K_p$ (1.5$\times$)} \\
        \midrule
        SPACE           & 96\% & 95\% & 100\% \\
        Control command & 98\% & 25\% & 0\% \\
        \bottomrule
    \end{tabular}
    \caption{Success rates on the PnP Box task under controller stiffness gains multiplied by different factors from training.}
    \label{tab:franka_gain_change}
    \vspace{-10pt}
\end{wraptable}
As shown in the table, increasing proportional gains by only $1.5\times$ drops the policy success rate from 98\% to 0\% for the control command policy. Qualitatively, the robot over-reaches the box and fails to grasp it in all cases. Similarly, reducing proportional gains by 0.5$\times$ also reduces the success rate to $25\%$. The robot fails to place the box on the desk by being stuck on the edge of the desk.
Meanwhile, \framename achieves similar success rates to the execution under the original gains, demonstrating the capability of dynamics adaptation. 
This is because \algname adapts to the different controller gain settings from calibration and online update steps.
% Thanks to calibration and online update steps, \algname can adapt to different controller gain settings when computing control commands that achieve the predicted Cartesian state delta.
% This is because of the adaptation capability of \algname, which adjusts control commands to achieve the predicted Cartesian state delta by considering the current controller gain settings.

% %===============================================================================

\section{Conclusion}
\label{sec:conclusion}
In this work, we introduced \framename, a framework that consists of a Cartesian state delta policy and \algname, which can effectively leverage datasets collected in different embodiments and robot hardware units.
% Compared to conventional policies that predict command input to the underlying controller, the Cartesian state delta policy predicts the recorded end-effector displacement, which is agnostic to varying dynamics across robots.
Compared to conventional policies that predict command input to the underlying controller, the Cartesian state delta policy predicts the recorded end-effector displacement, decoupling actions from robot-specific dynamics.
% Compared to conventional policies that predict command input to the underlying controller, the Cartesian state delta policy predicts the recorded end-effector displacement, decoupling actions from varying dynamics across robots.
% Compared to conventional policies that predict command input to the underlying controller, the Cartesian state delta policy predicts the recorded end-effector displacement, which is decoupled from the underlying mapping between command and robot motion that varies across robots.
\algname then outputs robot commands that achieve the predicted Cartesian state delta in the target robot.
By addressing discrepancies in commands across multiple robots, \framename improves policy performance for training across different embodiments and different hardware units. 
Furthermore, \framename remains robust under varying environment dynamics such as control Hz, payload, and controller gain variations.
This demonstrates that \framename enables reliable learning and deployment of policy across multiple robots.

\section{Limitations and Future Works}
\textbf{Force-aware manipulation.~}
While Cartesian state delta is a generalizable modality across different robot dynamics, it may not accurately reflect a force applied to an object when the same displacement can be achieved by applying different forces. 
% While Cartesian state delta is a generalizable modality across different robot dynamics, it may not accurately reflect a force applied to an object, especially in cases when the same displacement can be achieved by applying different forces. 
% For example, if the robot's hand is pushing the ground, no matter how hard it pushes, the achieved state delta will be zero.
To address this, future work could explore predicting the desired force for policy as in addition to the desired Cartesian state delta. 
% Still, this requires train data to contain force information.

\textbf{Different control modality conversion.~} In this work, Cartesian state delta is converted to Cartesian delta control command since they share the same modality. 
Deploying Cartesian state delta policy in robots that do not support Cartesian delta control commands will require converting Cartesian state delta to different control modalities (e.g., joint), which is not yet explored in this work.
% While \framename can learn from robots that do not support Cartesian delta control commands,
% deploying Cartesian state delta policy in such robots will require converting Cartesian state delta to different control modalities (e.g., joint), which is not yet explored in this work.
% (See Appendix~\ref{app:adapter_ablation} for initial attempts).

% ==== REMOVED TO FIT 8 PAGES ===
% \paragraph{Frame consistency}
% Cartesian state delta may be misaligned between robots using different frame conventions. For example, one could define robot end-effector pose in the base frame or the tool frame (end-effector frame). While not common, mapping of Cartesian coordinates to the actual robot movement could differ between embodiments. Training Cartesian state delta policy requires pre-processing of data if frame consistency does not meet.
%===============================================================================
% Force
% Velocity
% Joint space control
% Addressing these will be future work
% \paragraph{Multi robot evaluation}
% In this work, while we utilize UR5 and human data for experiments, we evaluate the resulting policy only in Franka Research 3 robot.

\clearpage
% The acknowledgments are automatically included only in the final and preprint versions of the paper.
\acknowledgments{
The authors thank Seonghyeon Ye, Dongkyu Shin, and Jaehwi Song for providing helpful comments for improving the work.
% If a paper is accepted, the final camera-ready version will (and probably should) include acknowledgments. All acknowledgments go at the end of the paper, including thanks to reviewers who gave useful comments, to colleagues who contributed to the ideas, and to funding agencies and corporate sponsors that provided financial support.
}

%===============================================================================

\bibliography{main}  % .bib
% Simulation result goes to Appendix

\newpage
\appendix
\begin{center}{\bf {\LARGE Appendix}}
\end{center}

\section{Pseudo Code}
\label{app:pseudo_code}
% \subsection{Action Adapter}
\begin{algorithm}[H]
\caption{SPACE rollout with \algname}
\label{alg:action_adapter}
\begin{algorithmic}[1]
\State \textbf{Given:} Cartesian state delta policy $\pi_\theta$
\State \textbf{Define:} Action adapter $\hat{u}=W_0\Delta p+b_0$ and its learning rate $\mu$
% \State \textbf{Define:} Action adapter $\adapter_\phi(\Delta p_t)=W_0\Delta p_+b_0$ and its learning rate $\mu$
\Statex
\vspace{-0.5em}
\State Collect $M$ calibration trajectories with length $K$, $\mathcal{D}_{\text{cal}}=\{\{p_0,u_0,\dots,p_K,u_K\}\}_{i=1}^{M}$
\State Initialize $W_0$ and $b_0$ by linear regression on $\mathcal{D}_{\text{cal}}$ using Equation~\eqref{eq:adapter_linear_regression}
\Statex
\vspace{-0.5em}
\For{$t \in [0, \dots, T-1]$}
    % \State Observe $o_t$ and $p_t$
    \State Predict target Cartesian state delta from $\Delta p_t^{\text{target}} \sim \pi_\theta(\cdot \mid o_t)$
    \State Run \algname $\hat{u}_t=W_t\Delta p_t^{\text{target}}+b_t$
    \State Execute $\hat{u}_t$ and observe next robot pose $p_{t+1}$
    \State Compute prediction error $e_t=W_t\Delta p_t^{\text{obs}}+b_t-\hat{u}_t$ for $\Delta p_t^{\text{obs}}=p_{t+1}-p_t$
    \State Update $W_{t+1} \leftarrow W_t-\mu e_t(\Delta p_t^{\text{obs}})^\top$, $b_{t+1} \leftarrow b_t-\mu e_t$
\EndFor
\end{algorithmic}
\end{algorithm}

\section{Experiment Details}
\label{app:exp_details}
\subsection{Action Adapter Calibration}
We collect calibration data $D_\text{cal}$ to train \algname. 
% To collect $\mathcal{D}_{\text{cal}}$ for training \algname, 
We execute a random scripted policy $\pi_{\text{rand}}$ for $M=10$ calibration trajectories, each with horizon $K=50$, yielding up to $500$ calibration steps. The policy generates Cartesian delta commands in a 6D end-effector space.
Specifically, we sample a random vector $\epsilon \sim \mathcal{N}(0, I)$ and multiply the step size uniformly from $[0.002, 0.01]$ for position and  $[0.005, 0.02]$ for orientation. The policy maintains a random direction vector and updates it every 10 steps by mixing $70\%$ of the previous direction with $30\%$ Gaussian noise. 
% The generated trajectory is perturbed with Gaussian noise of standard deviation $0.001$ for position and $0.0005$ for orientation, clipped within the workspace, and converted into desired Cartesian deltas. 
% These deltas are further clipped to maximum per-step magnitudes of $0.075$ m for position and $0.15$ rad for orientation before being sent to the robot. 
We record the achieved Cartesian state delta and the commanded action, forming calibration pairs $(\Delta p, u)$ used to fit \algname by ordinary least squares as follows:
\begin{equation}
\min_{W_0, b_0} \;
\sum_{(\Delta p, u) \in \mathcal{D}_{\text{cal}}}
\left\| W_0 \Delta p + b_0 - u \right\|_2^2.
\end{equation}
The entire process takes approximately 1 minute on the FR3 robot, and \algname fits in negligible time using a closed-form solution.
% \[
% \Theta^* = 
% \begin{bmatrix}
% W_0 & b_0
% \end{bmatrix}
% =
% U^\top \bar{P}
% \left(\bar{P}^\top \bar{P}\right)^{-1},
% \]
% where $\bar{P} = [\Delta p_1, 1; \ldots; \Delta p_N, 1]$ denotes the augmented matrix of achieved Cartesian state deltas and $U = [u_1; \ldots; u_N]$ denotes the corresponding commanded actions.

% To collect $D_\text{cal}$ used for \algname training, we use random scripted policy $\pi_rand$ and gather $M=10$ trajectories of length $K=50$. $\pi_\text{rand}$ is executing a sequence of random actions, constructed by random noise sampled from a Gaussian.
% Specifically, we sample a random vector $\epsilon \sim \mathcal{N}(0, I)$ and execute it for 10 steps to generate a motion in a specific direction. This vector is re-sampled every 10 steps and 

% === ORIGINAL ===
% To collect $\mathcal{D}_{\text{cal}}$, we roll out a random scripted policy $\pi_\text{rand}$ and gather $M$ trajectories of length $K$, i.e., $\mathcal{D}_\text{cal} = \{\tau^i_\text{rand}\}_{i=1}^{M}$ where each $\tau^i_\text{rand} = \{p_0, u_0, \dots, p_K, u_K\}$.

\subsection{Cross-Embodiment Experiment}
% We evaluate whether \framename more effectively leverages datasets from different embodiments than the control command policy.
% We evaluate whether the Cartesian state delta policy, executed with \algname, more effectively leverages datasets from different embodiments than the control command policy.

\textbf{Transfer from UR5 to FR3.~}
We use 250 demonstrations from the UR5 robot in the Berkeley UR5 Demonstration dataset~\citep{BerkeleyUR5} to train policies for the FR3 robot on four tasks: \emph{Bottle}, \emph{Cup}, \emph{Doll}, and \emph{Cloth}. As shown in Figure~\ref{fig:cross_embodiment_env}, the robot places a bottle into the pot (Bottle), stacks a blue cup on top of the brown cup (Cup), places a doll from the red bowl into the white bowl (Doll), and sweeps the cloth to the left side of the table (Cloth). 
% For the Bottle, Cup, and Doll tasks, we collect 20 FR3 demonstrations and co-train with the 250 UR5 demonstrations with balanced batch. For the Cloth task, we use only UR5 data and evaluate zero-shot transfer to the FR3.
For the Bottle, Cup, and Doll tasks, we collect 20 FR3 demonstrations and co-train with the 250 UR5 demonstrations. 
During FR3 data collection, the object is randomly placed on the desk, and the evaluation is conducted at fixed locations on the desk. 
% We refer to Figure~\ref{} for the object placement locations during evaluations. 
During training, we apply dataset balancing so that each batch has an equal amount of Franka data and UR5 data. This is realized by sampling each data point within the minibatch with uniform probability from the Franka or the UR5 robot. Also, we mask the UR5 proprioceptive state to be zero when co-training with Franka, since absolute proprioceptive states may not be shared between different embodiments.
% We refer to Figure~\ref{} for placement of object
For the Cloth task, we use only UR5 data and evaluate zero-shot transfer to the FR3.

% \textbf{Transfer from }
\textbf{Transfer from human hand-held gripper data.~}
We further test whether \framename is also effective for utilizing hand-held gripper data, specifically from the FastUMI dataset~\citep{zhaxizhuoma2025fastumi} (which we refer to as ``UMI data'' for simplicity).
Since UMI data lacks a control command, we naturally adopt the Cartesian state delta as an action space, which is calculated from the recorded robot pose in FastUMI.
% We study co-training with UMI and robot data to enable execution in deployment settings beyond those captured by UMI data alone.
We study co-training with UMI and robot data to enable execution in different domains beyond human data collection.
We use the ``PnP marker'' task, which comprises 550 UMI trajectories in which a marker is grasped and placed into a bowl, and co-train with 10 FR3 robot trajectories collected for the same task.
During evaluation, we vary the location of the marker in 8 different locations and fix the location of the bowl. The first 4 actions among an action chunk of length 16 are executed. We found that executing the full 16 actions in an open-loop leads to lower performance (5\% success rate for both SPACE and control command policy), and executing only one action out of 16 action chunks leads to a significant pause for control command policy.
We use zero vectors for proprioceptive input for the policy.

\subsection{Cross-Hardware Experiment}
We test whether \framename improves performance when training and deploying across different hardware units of the same embodiment.

\textbf{Transferring a policy across hardware units.~}
To evaluate the impact of these dynamics differences on policy performance, we compare \framename and the control command policy, trained using 50 demonstrations of the PnP Box task where the robot moves a box from the floor onto the desk. The resulting policies are evaluated on both the original robot used for data collection (Seen) and a new hardware unit (Unseen).
We use two Franka Research 3 robots for this purpose, one of which was manufactured in 2022 and the other in 2024. We use the same controller based on DROID and Polymetis~\citep{khazatsky2024droid}, and also set the inertia matrices and other gravity compensation variables equally to ensure that dynamics differences do not originate from those factors.

\textbf{Learning from multi-hardware data.~}
We use the DROID dataset~\citep{khazatsky2024droid}, a large-scale open-source dataset of FR3 robot demonstrations collected across multiple labs and hardware units. 
% Since DROID aggregates data from many hardware units, the control commands reflect a mixture of dynamics, providing a noisier supervision signal for the policy.
To reduce training cost, we filter DROID trajectories whose instructions contain ``marker", yielding 6614 trajectories with 1.4M samples, on which we fine-tune $\pi_{0.5}$ for 5 epochs with a batch size of 64.
The learned policy is evaluated on two tasks, ``Marker to Cup'' and ``Remove Marker from Cup'', in which the robot moves a marker into a cup on the tray and moves a marker inside the cup onto the tray, respectively. 
During evaluation, we fix the location of the marker and the cup.

% \subsection{Dynamics Shift Experiment}
% \textbf{Increasing control frequency}
% Specifically, we take the $\pi_{0.5}$ model trained on the PnP Box task in Section~\ref{sec:cross_hardware} and increase the execution frequency from 15Hz (used during data collection) to 30Hz.

\section{Extended Results}
\label{app:extended_results}
\subsection{Replay Tracking Test}
\label{app:tracking_test}
% In your preamble:
% \usepackage{wrapfig}
% \usepackage{booktabs}
% UR5 figure
\begin{table}[H]
\centering
\small
\setlength{\tabcolsep}{4pt}
\renewcommand{\arraystretch}{1.1}
\begin{tabular}{llcc}
\toprule
\textbf{Task} & \textbf{Replay method} & \textbf{Pos Err (mm)} & \textbf{Rot Err ($^\circ$)} \\
\midrule
\multirow{2}{*}{Bottle}
& Control command                 & 123.7 & 33.57 \\
& Cartesian state delta + \algname & 15.2 & 8.52 \\
\midrule
\multirow{2}{*}{Cup}
& Control command                 & 118.5 & 9.49 \\
& Cartesian state delta + \algname & 15.0 & 1.61 \\
\midrule
\multirow{2}{*}{Doll}
& Control command                 & 116.0 & 7.96 \\
& Cartesian state delta + \algname & 17.4 & 1.96 \\
\midrule
\multirow{2}{*}{Cloth}
& Control command                 & 169.7 & 15.04 \\
& Cartesian state delta + \algname & 19.8 & 2.72 \\
\bottomrule
\end{tabular}
\vspace{4pt}
\caption{Tracking error of replaying UR5 trajectories in Franka across four tasks. }
\vspace{-4pt}
\label{tab:ur5_replay_tracking_error}
\end{table}
% \begin{table}[H]
% \centering
% \small
% \setlength{\tabcolsep}{4pt}
% \renewcommand{\arraystretch}{1.1}
% \begin{tabular}{lcc}
% \toprule
% \textbf{Replay} & \textbf{Pos Err (mm) $\downarrow$} & \textbf{Rot Err ($^\circ$) $\downarrow$} \\
% \midrule
% Cartesian delta state               & 0.0 $\pm$ 0.0 & 0.00 $\pm$ 0.00 \\
% \quad + \algname                    & 0.0 $\pm$ 0.0 & 0.00 $\pm$ 0.00 \\
% % \quad + Aggregation                 & 0.0 $\pm$ 0.0 & 0.00 $\pm$ 0.00 \\
% Control command                     & 0.0 $\pm$ 0.0 & 0.00 $\pm$ 0.00 \\
% \bottomrule
% \end{tabular}
% \vspace{4pt}
% \caption{\textbf{Tracking error of replaying UR5 trajectory in Franka.}}
% \label{tab:ur5_replay_tracking_error}
% \end{table}
% \begin{wraptable}{r}{0.55\linewidth}
% \centering
% \small
% \setlength{\tabcolsep}{4pt}
% \renewcommand{\arraystretch}{1.1}
% \begin{tabular}{lcc}
% \toprule
% \textbf{Method} & \textbf{Pos. (mm) $\downarrow$} & \textbf{Rot. ($^\circ$) $\downarrow$} \\
% \midrule
% Cartesian delta state               & 0.0 $\pm$ 0.0 & 0.00 $\pm$ 0.00 \\
% \quad + \algname                     & 0.0 $\pm$ 0.0 & 0.00 $\pm$ 0.00 \\
% \quad + Aggregation                 & 0.0 $\pm$ 0.0 & 0.00 $\pm$ 0.00 \\
% Control command                     & 0.0 $\pm$ 0.0 & 0.00 $\pm$ 0.00 \\
% \bottomrule
% \end{tabular}
% \caption{\textbf{Tracking error of replaying UR5 trajectory in Franka.}}
% \label{tab:ur5_replay_tracking_error}
% \end{wraptable}
\textbf{Transfer from UR5 to FR3.~}
In this section, we show that discrepancies in control command between UR5 and Franka Research 3 (FR3) robot by replaying UR5 control command in FR3 and comparing the replayed trajectories to the original trajectories. 
Table~\ref{tab:ur5_replay_tracking_error} displays the replay tracking error in Cartesian coordinates (Pos) and orientation (Rot). As shown in Table~\ref{tab:ur5_replay_tracking_error}, replaying the UR5 control command in FR3 leads to a large tracking error. Meanwhile, replaying Cartesian state delta using \algname reduces tracking error by not relying on the original UR5 control commands. This double-confirms that  executing Cartesian state delta with \algname makes robot data shareable across different embodiments.

\subsection{Action Adapter Ablation}
\label{app:adapter_ablation}
\begin{wraptable}{r}{0.45\linewidth}
\centering
\small
\setlength{\tabcolsep}{4pt}
\renewcommand{\arraystretch}{1.1}
\begin{tabular}{lc}
\toprule
\textbf{PnP Box} & \textbf{Success Rate} \\
\midrule
\textbf{Ours}: Offline + online & \textbf{96\%} \\
Offline only & 15\% \\
Online only & 0\% \\
Online only (continuous) & 30 \% \\
% \textbf{Ours}: Offline + online (cont.) & 90\% \\
\bottomrule
\end{tabular}
\vspace{2mm}
\caption{Offline and online update ablation of \algname on the PnP Box task. The success rates are measured across 20 rollouts, except for ours, which uses 50.}
\label{tab:ablation_online_offline_update}
\end{wraptable}
\textbf{Offline and online update ablations.~}
We ablate the components of \algname to assess their contribution to performance. 
Specifically, we test whether (i) continuously updating \algname online and (ii) initializing \algname from calibration trajectories are both necessary. To this end, we consider two variants: (1) \textbf{Offline only}, which initializes $W_0$ and $b_0$ from $\mathcal{D}_\text{cal}$ but does not update them online, and (2) \textbf{Online only}, which skips calibration by initializing $W_0=I$ and $b_0=\mathbf{0}$. We additionally consider \textbf{Online only (continuous)}, which initializes the same way as Online only but continuously updates \algname across the 20 rollouts (to which the success rate is also measured). This yields roughly 3000 steps of LMS update data, substantially more than the 500 steps in $\mathcal{D}_\text{cal}$. The resulting success rates on the PnP Box task are reported in Table~\ref{tab:ablation_online_offline_update}.

As shown in Table~\ref{tab:ablation_online_offline_update}, removing the online update (Offline only) leads to a significant drop in success rate, showing that the initial parameters of \algname become inaccurate as the robot's pose changes during rollout. In contrast, the Online only variant achieves a 0\% success rate, demonstrating the importance of offline initialization. Although Online only (continuous) improves the success rate to 30\% by continuously adapting \algname, it still lags far behind our full method that combines offline initialization with online updates. This is because LMS-based online updates are designed to track recent dynamics rather than to aggregate information across many rollouts, so they cannot effectively exploit the larger amount of accumulated data. 
% We refer readers to Appendix~\ref{sec:app_extended_results} for additional ablations of \algname, including the choice of initial calibration trajectories, conversion of Cartesian state delta to joint control commands, and comparisons with gain-tuning baselines.

\begin{figure}[ht]
    \centering
    % \vspace{-5pt}
    % \includegraphics[width=0.7\textwidth]{figures/figure_cross_embodiment_environment.pdf}
    \includegraphics[width=0.88\textwidth]{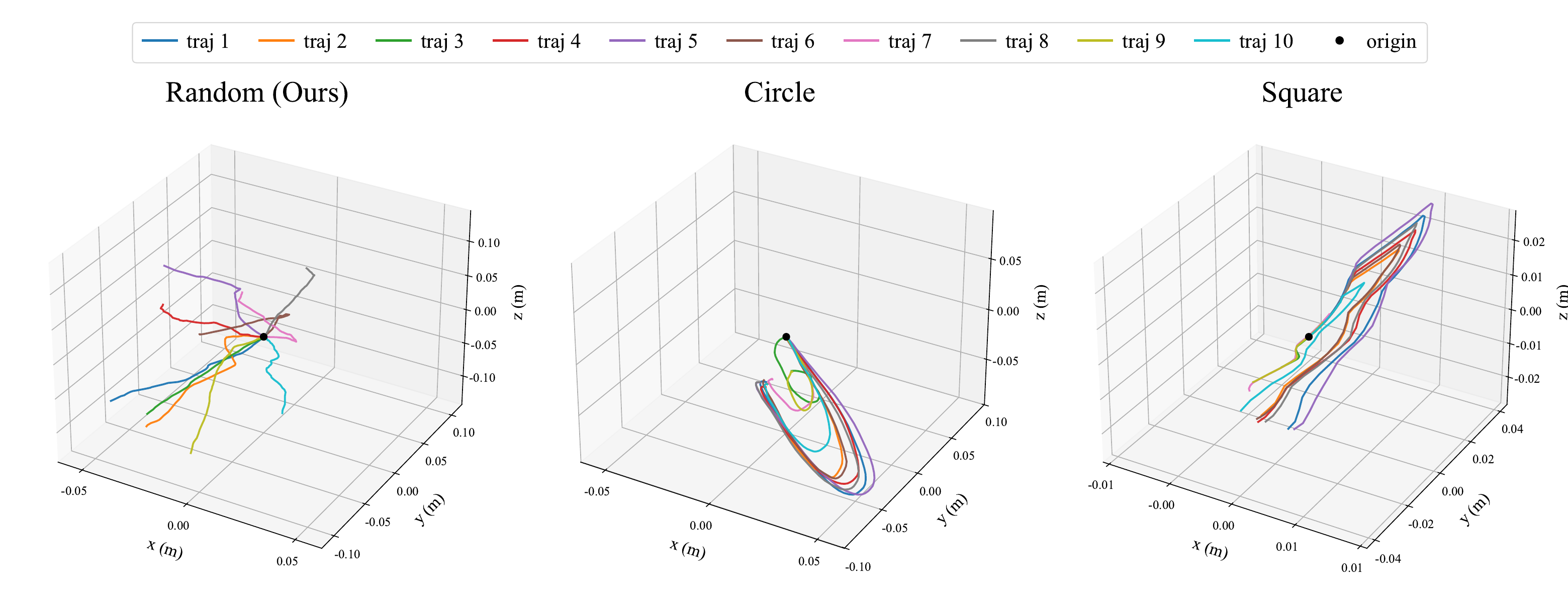}
    \caption{Different trajectory for $D_\text{cal}$ used for ablation with circle and square shapes in addition to our original choice, random trajectories.}
    \label{fig:calib_traj}
    \vspace{-8pt}
\end{figure}

\textbf{Calibration trajectories.~}
We also ablate how different choices for initial calibration trajectories affect the \algname performance. To test this, we attempt new calibration trajectories named \textit{Circle} and \textit{Square}, in which calibration trajectories draw a circle and a square shape. Then, we multiply a random constant sampled uniformly from $[0.2, 4.0]$ to vary the scale of the trajectory for increasing diversity, as shown in Figure~\ref{fig:calib_traj}.
% Details on shape generation.
Then, we replay the trajectory used for the tracking test in Figure~\ref{fig:replay_error}, using Cartesian state delta but with \algname learned from those different calibration trajectories. 

% \begin{table}[H]
% \centering
% \small
% \setlength{\tabcolsep}{4pt}
% \renewcommand{\arraystretch}{1.1}
% \begin{tabular}{lc}
% \toprule
% \textbf{PnP Box} & \textbf{Success Rate} $\uparrow$ \\
% \midrule
% \textbf{Ours}: Random & \textbf{90\%} \\
% Wave & 0\% \\
% Circle & 0\% \\
% \bottomrule
% \vspace{2pt}
% \end{tabular}
% \caption{\textbf{Initial calibration trajectory.}}
% \label{tab:ablation_cal_trajectory}
% \end{table}
% \begin{table}[H]
% \centering
% \small
% \setlength{\tabcolsep}{4pt}
% \renewcommand{\arraystretch}{1.1}
% \begin{tabular}{lccc}
% \toprule
% \textbf{Calibration} 
% & \textbf{Pos Err (mm)} 
% & \textbf{Rot Err ($^\circ$)}
% & \textbf{Success Rate} \\
% \midrule
% \textbf{Ours}: Random & 0.0 $\pm$ 0.0 & 0.00 $\pm$ 0.00 & \textbf{90\%} \\
% Circle                  & 0.0 $\pm$ 0.0 & 0.00 $\pm$ 0.00 & 0\% \\
% Square                & 0.0 $\pm$ 0.0 & 0.00 $\pm$ 0.00 & 0\% \\
% \bottomrule
% \end{tabular}
% \vspace{2pt}
% \caption{\textbf{Initial calibration trajectory.}}
% \label{tab:ablation_cal_trajectory}
% \end{table}
% Robot 1 (data collection) & $6.39 \stdv{0.00}$ & $1.36 \stdv{0.01}$ \\

% \begin{table}[H]
% \centering
% \small
% \setlength{\tabcolsep}{4pt}
% \renewcommand{\arraystretch}{1.1}
% \begin{tabular}{lcc}
% \toprule
% \textbf{Calibration} 
% & \textbf{Pos Err (mm)} 
% & \textbf{Rot Err ($^\circ$)} \\
% \midrule
% \textbf{Ours}: Random & 6.39 & 1.36 \\
% Circle                & 66.96 & 6.87 \\
% Square                & 143.94 & 10.30 \\
% \bottomrule
% \end{tabular}
% \vspace{4pt}
% \caption{\textbf{Initial calibration trajectory.}}
% \label{tab:ablation_cal_trajectory}
% \end{table}
\begin{wraptable}{r}{0.48\textwidth}
\centering
% \vspace{-10pt}
\small
\setlength{\tabcolsep}{4pt}
\renewcommand{\arraystretch}{1.1}
\begin{tabular}{lcc}
\toprule
\textbf{Calibration} 
& \textbf{Pos Err (mm)} 
& \textbf{Rot Err ($^\circ$)} \\
\midrule
\textbf{Ours}: Random & 6.39 & 1.36 \\
Circle                & 66.96 & 6.87 \\
Square                & 143.94 & 10.30 \\
\bottomrule
\end{tabular}
\vspace{2pt}
\caption{Replay tracking errors of \algname across different initial calibration trajectory choices for collecting $D_\text{cal}$.}
\label{tab:ablation_cal_trajectory}
\vspace{-10pt}
\end{wraptable}

As shown in the table, using calibration trajectories with circle and square shapes increases tracking error significantly. We hypothesize that this is due to decreased diversity in calibration trajectories, which might overfit \algname to a specific pattern in the calibration trajectories. In contrast, generating calibration trajectories using random actions allows the robot to visit diverse states, leading to less overfitting to less diverse poses.

\subsection{Baselines of Action Adapter}
% \begin{table}[H]
% \centering
% \small
% \setlength{\tabcolsep}{4pt}
% \renewcommand{\arraystretch}{1.1}
% \begin{tabular}{lcc}
% \toprule
% \textbf{Method} & \textbf{PnP Box} & \textbf{Stack Cube} \\
% \midrule
% \textbf{Ours} & \textbf{90\%} & \textbf{--} \\
% Gain tuning & 0\% & -- \\
% Delta aggregation & 0\% & -- \\
% \bottomrule
% \end{tabular}
% \vspace{2pt}
% \caption{\textbf{Comparison to baselines of \algname.}.}
% \label{tab:baseline_comparison}
% \end{table}
\begin{table}[H]
\centering
\small
\setlength{\tabcolsep}{4pt}
\renewcommand{\arraystretch}{1.1}
\begin{tabular}{lccc}
\toprule
\textbf{Method} 
& \textbf{Pos Err (mm)} 
& \textbf{Rot Err ($^\circ$)}
& \textbf{Success Rate (PnP Box)} \\
\midrule
\textbf{\algname}       & 14.6 & 1.4 & \textbf{94\%} \\
Gain tuning         & 29.9 & 2.9 & 20\% \\
Delta accumulation   & 40.1 & 2.4 & 45\% \\
\bottomrule
\end{tabular}
\vspace{4pt}
\caption{Comparison to alternatives of \algname. Gain tuning searches for controller gains that can track the desired Cartesian state delta. Delta accumulation accumulates the predicted Cartesian state delta to compute an absolute pose for tracking.}
\label{tab:baseline_comparison}
\vspace{-15pt}
\end{table}
In this section, we explore the alternatives of \algname in converting the Cartesian state delta to robot-specific commands that realize it. 

\textbf{Gain tuning.~} We attempt gain-tuning so that the robot well-tracks the Cartesian state delta without under-reaching. Concretely, we search for the combinations of increasing end-effector proportional gains by $\times [2, 4, 6, 8]$ with damping gains by $\times[1,2,3,4]$ and choose the ones with the lowest tracking error on \algname calibration trajectories, resulting in $\times6$ proportional gains and $\times$2 damping gains. Then, we additionally vary the proportional and damping gains in the joint space for the best end-effector space gains identified, resulting in increased $\times$8 joint proportional gains with joint damping gain values retained. 
Using these tuned gains, we conduct a replay test using the same trajectory used in Table~\ref{tab:ablation_cal_trajectory} and report the tracking error in Table~\ref{tab:baseline_comparison}. We additionally report the success rates of directly executing the Cartesian state delta policy under those tuned gains (measured over 20 rollouts). As shown in Table~\ref{tab:baseline_comparison}, the Cartesian state delta policy under those tuned gains still underperforms \framename using \algname. We suppose this is due to the absence of online update in the gain-tuning baseline, given that removing online update in \algname results in a significant performance drop~(Table~\ref{tab:ablation_online_offline_update}). 
% Also, unless one searches for very extensive available gains using a vast amount of resources, fine-grained loss of gain adjusting is compared to linear regression in \algname. Furthermore, it cannot be applied to closed-source robot controllers.
Moreover, unless one searches over an extensive gain space at considerable computational cost, gain tuning offers less fine-grained control compared to \algname, and it is inapplicable to closed-source robot controllers altogether.

\textbf{Delta Accumulation.~}
Additionally, we consider the delta accumulation technique, where the sequence of Cartesian state deltas $(\Delta p_t^{\text{target}}, \dots, \Delta p_{t+H-1}^{\text{target}}) \sim \pi_\theta$ predicted by the policy action chunk is accumulated to construct absolute target poses.
By accumulation, the controller tracks $p_{t+k}^{\text{target}} = p_t + \sum_{j=0}^{k} \Delta p_{t+j}^{\text{target}}$ at rollout timestep $t+k$ for $k=1,\dots,H$ when policy action chunk length is $H$.
Compared to commanding delta, commanding absolute poses does not aggregate under-reaching error from individual steps and is thus expected to perform better at tracking.
However, as shown in Table~\ref{tab:baseline_comparison}, the accumulation heuristic performs poorly compared to using \algname. This is because using an absolute target does not fully reduce the tracking error, and the absence of an online update makes it unable to reflect the most recent robot dynamics.
\end{document}